\documentclass[10pt,journal,compsoc]{IEEEtran}
\usepackage{comment}
\usepackage{rotating}
\usepackage{makecell, multirow}
\setcellgapes{3pt}
\usepackage{amssymb}
\usepackage{amsmath}
\usepackage{pifont}
\usepackage{threeparttable} 
\usepackage{bm}
\usepackage{booktabs}
\usepackage[table,xcdraw]{xcolor}
\usepackage[utf8]{inputenc}
\usepackage{diagbox}
\usepackage{url}
\usepackage{threeparttable} 
\usepackage{bm}
\usepackage{tikz}
\usepackage{subfig}
\usepackage{graphicx}
\usepackage[english]{babel}
\newcommand{\smark}{\textcolor{black}{\textbf{--}}}

\DeclareMathOperator*{\argmin}{arg\,min}

\ifCLASSOPTIONcompsoc
  \usepackage[nocompress]{cite}
\else
  \usepackage{cite}
\fi
\ifCLASSINFOpdf
\else
\fi
\begin{document}
%
\title{Improving the Reliability of Network Intrusion Detection Systems through Dataset Integration}

%
%
%
%

\author{Roberto~Magán-Carrión, 
        Daniel~Urda, 
        Ignacio~Diaz-Cano,
        and~Bernabé~Dorronsoro
\IEEEcompsocitemizethanks{\IEEEcompsocthanksitem R. Magán-Carrión is with Network Engineering \& Security Group (NESG), Dpt. of Signal Theory, Telematics \& Communications, School of Computer Engineering \& Telecommunications, University of Granada (Spain).\protect\\
E-mail: rmagan@ugr.es
\IEEEcompsocthanksitem D. Urda is with Grupo de Inteligencia Computacional Aplicada (GICAP), Departamento de Ingeniería Informática, Escuela Politécnica Superior, Universidad de Burgos (Spain).\protect\\
E-mail: durda@ubu.es
\IEEEcompsocthanksitem I. Díaz-Cano and B. Dorronsoro are with the Superior Engineering School of the University of Cadiz (Spain).\protect\\
E-mail: \{ignacio.diaz, bernabe.dorronsoro\}@uca.es
}
\thanks{Manuscript received October 26, 2021; revised ---.}}

%
%

\markboth{IEEE Transactions on Emerging Topics in Computing}%
{Magán-Carrión \MakeLowercase{\textit{et al.}}: Improving the Reliability of NIDSs through Dataset Integration}
%



\IEEEtitleabstractindextext{%
\begin{abstract}
This work presents Reliable-NIDS (R-NIDS), a novel methodology for Machine Learning (ML) based Network Intrusion Detection Systems (NIDSs) that allows ML models to work on integrated datasets, empowering the learning process with diverse information from different datasets. Therefore, R-NIDS targets the design of more robust models, that generalize better than traditional approaches.
We also propose a new dataset, called UNK21. It is built from three of the most well-known network datasets (UGR'16, USNW-NB15 and NLS-KDD), each one gathered from its own network environment, with different features and classes, by using a data aggregation approach present in R-NIDS. Following R-NIDS, in this work we propose to build two well-known ML models (a linear and a non-linear one) based on the information of three of the most common datasets in the literature for NIDS evaluation, those integrated in UNK21. The results that the proposed methodology offers show how these two ML models trained as a NIDS solution could benefit from this approach, being able to generalize better when training on the newly proposed UNK21 dataset. Furthermore, these results are carefully analyzed with statistical tools that provide high confidence on our conclusions.
\end{abstract}

}

\maketitle

\IEEEdisplaynontitleabstractindextext

%
\IEEEpeerreviewmaketitle

\IEEEraisesectionheading{\section{Introduction}\label{sec:introduction}}
Several technical reports announce the extraordinary growing of the devices connected to IP (Internet Protocol) networks nowadays and in a near future. For instance, the Cisco Annual Internet Report (2018–2023)~\cite{cisco_2020_report} forecasts that its number will exceed 29 billions, over three times the global population. This fact can be motivated by the advent of the real deployment of communications technologies like 5G, providing easy and affordable Internet access for a huge number of heterogeneous devices from IoT (Internet of Things) ecosystems. Such an \mbox{(inter-)} connectivity allows to develop new applications and to offer new services inconceivable before. However, both the security risk and the attack surface grow as the number of connected devices, communications and services increase. As a consequence, malicious actors have many ways and exposed entry points to perform and execute attacks. According to the ENISA Threat Landscape 2020 report (ETL)~\cite{ENISA-2020}, the sophistication of threat capabilities seriously increased in 2019, having detected over 200,000 daily new variants of malware targeting diverse objectives.

In this scenario, there is an major need for novel methods and techniques for an early attack detection, which is of upmost importance to reduce the impact of the attacks on the target system or communication network. These new solutions must be robust, resilient, and adaptable to highly dynamic scenarios. According to the ETL report~\cite{ENISA-2020}, the detection of an attack takes around 6 months in average thus, a short detection time reduces the attack's impact.

IDSs (Intrusion Detection Systems) have been widely used to address this problem in general, being the NIDSs (Network IDSs) used for monitoring and detecting malicious actions in communications networks. A large number of Machine Learning (ML)-based solutions have been proposed in the literature for NIDSs~\cite{CJ21,Zoppi2021,Magan2020,hajisalem_hybrid_2018,kabir_2018}. These methods are generally trained and validated on one single dataset. However, despite the fact that some of the existing datasets were carefully built with realistic network traffic and attacks, such kind of studies does not allow providing generic conclusions about the performance of the proposed methods on other network datasets. Indeed, the trustworthiness of a method that is trained and validated on a single dataset is seriously compromised, and this is a major issue nowadays in the literature of ML-based NIDS methods. Besides, this is a critical concern in the topic, because these methods must detect network attacks (ideally, including new attacks) in any scenario.

More reliable solutions could be designed if several network datasets are jointly considered during the training process of the ML model. In principle, this may allow capturing the fundamentals of heterogeneous attacks observed and/or artificially generated in different network environments where each single network dataset was created. Unfortunately, integrating data from several datasets is a difficult task in this problem, given that all existing ones in the literature present important differences. For instance they differ in the number of features, their values, the annotated attacks, whether they include timestamp or not in the observations, etc. This makes it unfeasible to apply a given solution on different datasets with no additional methods or techniques.

Besides, ML-based techniques typically require the normalization of input data to avoid any undesirable bias due to existing differences in the magnitudes of the variables' values. In this sense, samples fed to ML methods must be normalized in order to achieve an acceptable generalization performance under the assumption of both the development and test sets having the same distribution. However, this assumption does not always hold since the data used to train a ML model may differ from the one present in the testing scenario, a well-known issue in ML-based approaches commonly known as data shift~\cite{QSS09,Heiser_2019} (e.g., differences on the class labels definition ---shift on the output---, differences on the distribution of input features ---shift on the \mbox{input---)}. In fact, this issue is more likely to be present in network datasets for NIDSs since each of them are built on different network environments and under different experimental settings (e.g., real or synthetic network traffic, variety of attacks samples both real or manually generated, etc.). Consequently, one needs to solve the challenge of normalizing data belonging to different network datasets in order to put ML models that generalize well into production, in terms of accurately classifying the traffic of a real network.

Camacho \emph{et al.} proposed in~\cite{camacho_tackling_2014} a data aggregation technique for network datasets, called \emph{Feature as a Counter} (FaaC). It aggregates a batch of raw network traffic observations into one single sample by translating the dataset features into a new set of features that represent counters of the previous ones. FaaC was later successfully applied to NIDS in~\cite{Magan2020}, where authors used as a case of study the UGR'16 dataset~\cite{Macia-Fernandez2018}, a timestamp-based network dataset built from real network traffic. It is reasonable to think that time distribution (i.e., network flows within a certain time slot) might contain relevant information to detect attacks. However, its applicability to other well-known previous network datasets, such as USNW-NB15~\cite{moustafa_unsw_nb15_2015} or NLS-KDD~\cite{Tavallaee2009}, the latter with no timestamp information, is not a priory possible, or at least as the FaaC technique was conceived. 

In this sense, a novel methodology to apply FaaC to timestamp-less datasets was recently proposed in \cite{magan-carrion_assessing_2022}. The paper shows that there are no major differences between the performance of ML models trained on this and the timestamp-based versions of the UGR'16 dataset. Based on this result, we propose in this work a novel methodology that allows using and combining several network datasets, each one with their own set of input features, to train reliable ML-based NIDSs solutions that generalize better across different network environments, in contrast to the single dataset-based solution that is typically found in the literature. 
Additionally, this novel methodology allows normalizing the derived datasets in an easy and flexible way (no matter the number of samples, the values in their variables nor the network environment used to build the raw dataset), a key feature that makes feasible the use of the previously built ML models in a production scenario~\cite{magan-carrion_multivariate_2020}, in comparison to other existing ML-based NIDSs solutions. Our main hypothesis is that the resulting ML model developed using this novel methodology may benefit from the overall attacks information contained across different datasets, which gives hope to achieve ML-based NIDSs solutions that outperform classical approaches that are only trained in one single dataset. Therefore, this work supposes the first step towards the design of a new generation of reliable ML-based NIDSs that can perform accurately in generic network scenarios.

The main contributions of this work are summarized as follows:
\begin{itemize}
    \item The deployment of a complete methodology, Reliable-NIDS (R-NIDS), that allows building trustworthy ML models for NIDSs, as well as evaluating the results in a reliable way, including steps for (i) pre-processing and transforming raw network datasets, (ii) performing data integration of different network datasets, (iii) training robust ML models ---feature selection, hyper-parameter selection, evaluation strategy, performance metrics---, and (iv) testing statistical significance. Apart from that, R-NIDS methodology allows to design, implement and make fair and honest comparisons of ML-based solutions for NIDSs which is not commonly seen in the literature.
    \item The introduction of a novel data integration approach, that allows to jointly consider different network datasets through a row-wise aggregation step. Consequently, the new UNK21 dataset is proposed, combining network traffic from three previous and well-known network datasets: UGR'16, UNSW-NB15 and NSL-KDD.
    \item The implementation of a reliable ML-based solution using the new UNK21 dataset for NIDS, which is shown to considerably generalize better than previous existing ones that were based on one single network dataset. Particularly, this solution can be used in production in generic network scenarios thanks to the data aggregation and normalization approach presented in this paper. Furthermore, it is also capable of analyzing network communication flows regardless of the aggregation strategy employed for the FaaC technique, i.e., analyzing traffic within a certain frequency (timestamp-based) or within a set of a certain size of consecutive network flows (batch-based).
\end{itemize}

The rest of the paper is organized as follows. In Section~\ref{sec:background}, the most relevant works carried out to date regarding the evaluation of NIDS are studied. In Section~\ref{sec:datasets}, the datasets that have been considered in this study are introduced (UGR'16, USNW-NB15 and NSL-KDD). The proposed framework, called R-NIDS, is introduced and described in detail in Section~\ref{sec:methods}. Next, the experimental design employed in this work to develop and evaluate reliable ML-based solutions for NIDSs is presented in Section~\ref{sec:experiments}. Finally, Section~\ref{sec:results} presents and discusses the results obtained in the analysis, and the conclusions and future work are shown in Section~\ref{sec:conclusions}.

\section{Background}
\label{sec:background}

\begin{table*}[!t]
	\makegapedcells
	\centering
    \caption{Methodology comparison for NIDS evaluation.}
    \label{tab:methodology_comp}
    \resizebox{\linewidth}{!}{
    \begin{threeparttable}
    	\begin{tabular}{c c c c c c c c c}
    		\toprule
    		\multirow{2.3}{*}{\textbf{Work}} & \multirow{2.3}{*}{\textbf{Dataset}} & \multicolumn{7}{c}{\textbf{Methodology}} \\
    		
    		&                                                                                        & \textbf{FE} & \textbf{FS} & \textbf{DP} & \textbf{HS} & \textbf{ML} & \textbf{PM*} & \textbf{SSA}  \\
    		
    		\midrule
    		 Hajisalem \textit{et al.} \cite{hajisalem_hybrid_2018} & NSL-KDD, UNSW-NB15                 & \smark & \checkmark & \smark & \smark & other & A, TFR & \smark  \\ 
    		Kabir \textit{et al.} \cite{kabir_2018} & KDDCup'99                                   & \smark & \checkmark & \smark & \smark & classic & F1, R, P & \smark  \\
    		Sharafaldin \textit{et al.} \cite{sharafaldin_toward_2019} & CICIDS2017                     & \checkmark & \checkmark & \smark & \smark & classic & F1, R, P & \smark \\
    		Siddiqi  \textit{et al.}~\cite{Siddiqi2020} & NLS-KDD, ISCX12                         & \checkmark  & \checkmark & mean & \smark & classic & A, F1 & \smark   \\
    		Kumar  \textit{et al.}~\cite{Kumar2020} & UNSW-NB15                         & \checkmark  & \checkmark & normalization & \smark & other & A, R, P & \smark   \\
    		Tian  \textit{et al.}~\cite{Tian2020} & UNSW-NB15, NSL-KDD & \checkmark  & \checkmark & normalization & \smark & deep learning & R, O & \smark\\
    		Verma  \textit{et al.}~\cite{Verma2020} & CIDDS-001, UNSW-NB15, NSL-KDD & \checkmark  & \smark& \smark & \checkmark & classic, deep learning & A, TFP, AUC & Friedman, Nemenyi\\
    		Aleesa  \textit{et al.} \cite{Aleesa2021} & UNSW-NB15                         & \checkmark  & \checkmark & normalization & \checkmark & deep learning & A, P, ROC, AUC & \smark    \\
    		Toldinas  \textit{et al.}~\cite{Toldinas2021} & UNSW-NB15                         & \checkmark  & \checkmark & normalization & \smark & deep learning & A & \smark \\
    		Pooja \textit{et al.}~\cite{Pooja2021} & KDD'99, UNSW-NB15 & \smark & \smark & normalization & \smark & classic, deep learning & A & \smark\\
    		Zoppi  \textit{et al.}~\cite{Zoppi2021} & NLS-KDD, UGR16, CICIDS17, UNSW-NB15 & \checkmark  & \checkmark & normalization & \checkmark & deep learning & A, TFR, F1, R, P & \smark\\

            Magán-Carrión \textit{et al.}~\cite{Magan2020} & UGR'16                                    & \checkmark & \checkmark & normalization & \checkmark & classic & F1, R, P, AUC & Friedman, Wilcoxon  \\

    		\bottomrule
    	\end{tabular}
    	    \begin{tablenotes}
                \footnotesize
                \item[*] A (Accuracy), TFR (TP --True Positive Rate--, FP --False Positive Rate--, TN --True Negative Rate--, or  FN --False Negative Rate--), F1 (F1-score), R (Recall), P (Precision), ROC (Receiver Operating Characteristic curve), AUC (Area Under the Curve), O (Others) \\
            \end{tablenotes}
   \end{threeparttable}}
\end{table*}

Throughout this section, we will present and discuss some of the most interesting articles dealing with the detection of network attacks using ML-based techniques. Table~\ref{tab:methodology_comp} gives an overall picture and comparison of these works. It mainly shows information about key factors involved when implementing ML-based NIDSs solutions which are, among other, the datasets they work with and the stages that should be found in a work pipeline to train and test ML modes as recommended in~\cite{Magan2020}. They are the Feature Engineering (FE), the Data Pre-processing (DP), the Feature Selection (FS), the Hyper-parameters Selection (HS), the Machine Learning (ML) model and the Performance Metrics (PM) used. Additionally, we consider a new and important stage for the Significance Statistical Analysis (SSA). The `\checkmark' is used to point out that the corresponding work is considering a specific stage (existing or proposed by the authors) within their approach while the `\smark'  shows right the opposite (i.e., the stage is not included within the work or the authors miss or omit to mention it). Rest of values in the table are auto-explanatory.

A considerable amount of works have been recently published related to ML-based NIDSs. 
For instance, in~\cite{hajisalem_hybrid_2018} the authors proposed a NIDS solution for anomaly detection that relies on the joint use of swarm-intelligence-based optimization heuristics:  Artificial Fish Swarm (AFS) and Bee Colony Optimization (BCO). This hybrid algorithm, is able to detect anomalies using a reduced subset of characteristics. The proposed system implements an adequate feature selection procedure, though it does not introduce any of the other what we consider recommended steps. Finally, this study does not provide details about the method used, thus making it difficult to compare it against other solutions.

The authors in~\cite{kabir_2018} proposed a Least Squares Support Vector Machine (LS-SVM)-based NIDS. In this work, they used the Optimal Allocation technique to efficiently handle large datasets through the selection of the most representative samples, organized according to the number of observations and the confidence interval to be achieved. Despite the LS-SVM model is not usually used for intrusion detection, the authors considered the use of this technique interesting to define the entire population from a series of representative examples.

In ~\cite{sharafaldin_toward_2019}, the authors introduce and highlight the existing problems caused by the increasing number and variety of network attacks. They are in turn worsened due to the lack of appropriate data sets for training ML-based NIDSs. Therefore, they review a significant amount of network data sets from recent years, listing the problems that each one contains like: freshness, insufficient number of attacks, etc. As a result, they proposed a new dataset that solves, or at least mitigates the problems encountered. They tested several classical ML models to corroborate it. Similarly, authors in~\cite{Siddiqi2020} addressed this issue. In this case, they added the importance of making an effective feature selection for an efficient training and testing of ML models. For this purpose, they proposed to implement and evaluate the effects of normalizing samples before performing feature selection, concluding that this strategy provides better results than the previous existing ones.

In~\cite{Kumar2020} the authors proposed a system to detect five conventional attacks. The author built a new dataset that is afterwards compared with the USNW-NB15 dataset. In this sense, they employed the gain information technique over the set of features on the original USNW-NB15 dataset and used a misuse-based approach to create this new dataset. Such dataset was used to train ML models making them to perform better in terms of precision, attack detection rate, false positive rate, among other performance metrics.

The authors of~\cite{Tian2020} addressed a well-known problem in ML classification tasks: the overfitting problem. To overcome this problem, they proposed an approach based on improved Deep Belief Network (DBN) composed of multiple hidden layers and connections. Together with DBN, the dataset used was pre-processed using the Probabilistic Mass Function (PMF) and the Min-Max normalization technique that simplifies this procedure. Both techniques allowed to achieve a small improvement in the performance compared to other studies compared in this work. The authors considered all the stages we recommend here for a correct ML models training and testing except the hyper-parameter selection and the significance statistical analysis stages.

NIDS solutions are also being applied in IoT ecosystems, such as the one reported by the authors in~\cite{Verma2020}. This work proposes the use of classification algorithms to detect DoS attacks, one of the most damaging attacks for IoT platforms. The performance tests were made on ML popular datasets and using a Multi-Layer Perceptron (MLP). Although authors perform a hyper-parameter selection, other important phases, such as feature selection, were ignored. Additionally, they employed significance tests, specifically Friedman and Nemenyi, to statistically analyze the performance obtained, which should be a mandatory step in order to have confidence in the results obtained~\cite{stapor2021}.

A Deep Learning (DL)-based NIDS is featured in~\cite{Aleesa2021}. The authors proposed using a popular dataset, the USNW-NB15 to perform a binary and multi-class classification. In the stages of their methodology, they included all the steps suggested in our proposal except the significance statistical analysis. Subsequently, they compared the performance of the trained models with other published works, obtaining a performance rate close to 100\%. Similarly, the authors in~\cite{Toldinas2021} proposed a DL-based NIDS using the same dataset. This work tried to enhance state of the art NIDSs with use of image recognition techniques through different stages. Regarding the steps followed in their methodology, they did not mention anything about hyper-parameters selection. With respect to the dataset, they transformed it into four-channel images (red, green, blue, and alpha), which were subsequently used to train and validate the proposed model. Another DL-based system using bidirectional LSTM (Long Short-Term Memory) is introduced by the authors in~\cite{Pooja2021}. The experiments are carried out on classic datasets, and data normalization is only applied within the data pre-processing stage. The remaining stages were not contemplated by the authors, where only a performance study based on the considered activation function within the DL model was included.

In~\cite{Zoppi2021} the authors noted that supervised learning has been very effective for known threats, although it does not work well at detecting unknown attacks that are commonly called \textit{zero-day} attacks. In the study, the authors proposed a semi-supervised learning system, combining two layers to detect known and unknown threats, and addressing all the recommended steps for an adequate methodology in the study of network attacks, except the use of significance statistical tests. 

Therefore, considering Table~\ref{tab:methodology_comp} and the guidelines proposed in~\cite{Magan2020}, we can conclude that most of the works still fail in the use of HS and SSA stages, two of the most important steps for a reliable and honest design and comparison of NIDS solutions. It is also worth noting the recurrent use of the KDD-related datasets for validating NIDSs still nowadays, even though it is undoubtedly outdated (created almost 20 years ago in a field that significantly evolves within a year time). Moreover, authors tend to use the accuracy measure to validate their models, although it is widely known as a non-fair performance metric for highly imbalanced classification problems, such as the context of the problem addressed here. Finally, no works can still be found in the literature comprising the issue of making reliable ML-based NIDSs, aiming at performing well in generic network environments. In this work, we test our main hypothesis based on the idea that integrating network datasets would empower the learning process with diverse information from different datasets, at the same time that ML-based systems would be more adaptable in order to put them in production in generic network environments. 


\section{Network Datasets}
\label{sec:datasets}
Network security researchers continuously deal with a well-known issue concerning the existence of an adequate number of datasets to evaluate NIDS. For this reason, several efforts have been made in recent years in order to obtain good enough network datasets in some sense. For instance, some datasets stand out and differ from others in terms of the data format used (network flows or packets), the way in which the data is recorded (real or synthetic), the duration of the data recordings used in the dataset and their freshness~\cite{ring_survey_2019,Macia-Fernandez2018}.

Therefore, building the perfect dataset is not a trivial task, as stated in~\cite{ring_survey_2019,kenyon_are_2020}. This statement is mainly supported by two compelling reasons: 1) new attacks samples are continuously observed~\cite{enisa_etl_2017, enisa_etl_2018, ENISA-2020}, which makes it hard to have a permanently up-to-date dataset; and 2) the fact that specific applications context of ML-based models need specific datasets~\cite{Zoppi2021}.

Even in the case that no perfect dataset exists, in this study we hypothesized that ML-based solutions trained using a combination of datasets, could perform better than traditional solutions which involve only one dataset for training and testing. Moreover, this approach is more likely to be suitable for its implementation in real and generic production environments than others built using only one dataset.


In the following, we briefly introduce the three network datasets used in this work, to be afterwards combined into a single one called UNK21 (see Section~\ref{sec:experiments}). They are the UGR'16, USNW-NB15 and NSL-KDD, and were gathered in different network scenarios and also including a variety of attack samples generated with different tools. 

\subsection{UGR-16}

The UGR'16~\cite{Macia-Fernandez2018} dataset contains anonymized network traffic flows (NetFlow) collected during 4 months at the facilities of an Internet service provider (ISP) in Spain. Thus, the authors of this dataset have divided its content into two different parts: CAL, which refers to data obtained from March to June 2016, containing inbound and outbound ISP network traffic; and TEST, which corresponds to traffic obtained from July to August 2016, containing mostly attack streams that were either synthetically generated or found by detectors. 

In order to guarantee the correct functioning of the ISP and not adulterate the netflows with the synthetic attacks, they were launched under a controlled environment. Most of these attacks were created using updated tools that cover difficult to detect harmful attacks, like \textit{DoS (both low and high rate)}, \textit{Scan (Port Scanning)} or \textit{Botnet}. However, UGR'16 does not include only synthetic attacks, it also contains real attacks, detected and labeled by the authors. These detected attacks are \textit{UDP port scan}, \textit{SSH scan} and \textit{Spam} campaigns. Finally, and also of relevant importance, those streams generated by IP addresses from open blacklist sources were tagged as \textit{Blacklist}.

The dataset is composed by 23 files, one per week. From them, 17 were selected for CAL subset and the other 6 for TEST. The size of each file is around 14GB (compressed) and they can be downloaded in \textit{nfcapd} or CSV formats\footnote{The UGR'16 dataset: \url{https://nesg.ugr.es/nesg-ugr16/}}.

\begin{table}[t!]
\centering
\caption{Class distributions of the UGR'16 raw dataset.}
\label{tab:dist_clases_ugr16}
\begin{tabular}{c|c|c}
\toprule
{\textbf{\centering Class}} &
{\centering \textbf{raw obs.}} & {\centering \textbf{ \%}} \\ 
\midrule
 Background             & $\sim4,000M$   & $97.14$  \\
 Spam                   & $\sim78M$      & $1.96$   \\
 Blacklist              & $\sim18M$      & $0.46$   \\
 DoS                    & $\sim9M$       & $0.23$     \\
 Scan (Port Scanning)   & $\sim6M$       & $0.14$   \\
 Botnet                 & $\sim2M$       & $0.04$    \\
 UDP scan               & $\sim1M$       & $0.03$    \\
 SSH scan               & $64$           & $\sim0$  \\
 
\bottomrule
\end{tabular}
\end{table}

We summarize in Table \ref{tab:dist_clases_ugr16} the class distribution of the TEST part, together with the number of flows in each class. As it could be expected, the number of attack traffic flows is notably unbalanced with respect to normal traffic (\textit{Background}). Among the attacks, \textit{Spam} is the one with the highest number of flows, representing only $ 1.96 \% $ of the flows in TEST. In addition, TEST contains flows of all mentioned attacks, both synthetically generated and discovered from anomalous behaviors. For the mentioned reasons, we decided to use this part of the UGR'16 dataset in this work.

\subsection{USNW-NB15}

UNSW-NB15~\cite{moustafa_unsw_nb15_2015} is a public dataset that collects raw network traffic using the IXIA tool ``PerfectStorm'' in the CyberRange laboratory of the Australian Center for Cybersecurity (CASS). This set up consist of several network devices (firewalls and routers) connecting servers and hosts. In concrete, servers are in charge of generating both normal and malicious network traffic, from and to end devices, traversing a firewall.

After using the IXIA tool under this set-up, 47 characteristics were obtained from the data. The \textit{TCPdump} tool was installed on one of the routers to capture the PCAP files where a total of 2,540,044 records (traffic flows) were recorded an stored in four CSV files\footnote{The UNSW-NB15 dataset: https://researchdata.edu.au/unsw-nb15-dataset}. Nine different classes of attacks were identified: \textit{Fuzzers}, \textit{Analysis (Port Scanning)}, \textit{Backdoors}, \textit{DoS}, \textit{Exploits}, \textit{Generic}, \textit{Reconnaissance}, \textit{Shellcode}, and \textit{Worms}. For this study, the four CSV files of traffic flow records have been merged, and the number of samples that exist of each class have been counted, as shown in Table~\ref{tab:dist_clases_nb15}. Similarly to the UGR'16 dataset, there is a large percentage of background traffic in comparison to the other classes, followed by the the Generic attack. Other attack samples have a  lower presence in the datasets.

\begin{table}[t!]
\centering
\caption{Class distributions of the USNW-NB15 raw dataset.}
\label{tab:dist_clases_nb15}
\begin{tabular}{c|c|c}
\toprule
{\textbf{\centering Class}} &
{\centering \textbf{raw obs.}} & {\centering \textbf{ \%}} \\ 
\midrule
 Background      & $\sim2,21M$   & $87.35$   \\
 Generic         & $215,481$     & $8.48$   \\
 Exploits        & $44,525$      & $1.75$   \\
 Fuzzers         & $24,246$      & $0.95$    \\
 DoS             & $16,353$      & $0.63$    \\
 Reconnaissance  & $13,987$      & $0.54$     \\
 Shellcode       & $3.840$       & $0.14$   \\
 Analysis (Port Scanning) & $2,677$       & $0.09$   \\
 Backdoor        & $2,329$       & $0.07$   \\
 Worms           & $174$        & $\sim0$    \\
\bottomrule
\end{tabular}
\end{table}

\subsection{NLS-KDD}

NSL-KDD is a dataset built from the KDDCup'99 dataset. It was conceived to solve some of the well-kown flaws it raises e.g., the duplicity on the records~\cite{Tavallaee2009}. However, this new version of the KDD dataset still inherits some of the problems of the original one and it may not be considered a good representation of real traffic in communication networks. It is considered in this study to analyze its influence with respect to the other two involved datasets that are, in principle, more suitable according to their characteristics.

\begin{table}[!t]
\centering
\caption{Class distributions of the NSL-KDD raw dataset.}
\label{tab:dist_clases_kdd}
\begin{tabular}{c|c|c}
\toprule
{\textbf{\centering Class}} &
{\centering \textbf{raw obs.}} & {\centering \textbf{ \%}} \\ 
\midrule
 Background          & $67,343$ & $53.48$    \\
 DoS                 & $45,927$ & $36.47$   \\
 Probe (Port scanning)               & $11,656$ & $9.26$    \\
 R2L                 & $942$    & $0.75$      \\
 U2R                 & $52$     & $0.04$   \\
\bottomrule
\end{tabular}
\end{table}

The NLS-KDD dataset comprises two CSV files\footnote{The NSL-KDD dataset: https://www.unb.ca/cic/datasets/nsl.html} for training and testing purposes. In our case, we have merged the two files into only one. Table~\ref{tab:dist_clases_kdd} shows the number of flows of each class category. \textit{Probe (Port Scanning)} class comprises the following kind of attacks: \textit{psweep, nmap, portsweep, satan, saint, mscan}. Similarly, for the \textit{DoS} class, it comprises \textit{back, land, neptune, pod, smurf, teardrop, apache2, mailbomb, udpstorm, processtable} attacks. \textit{R2L (Remote to Local)} attacks which tries to get local access to remote target by exploiting certain vulnerabilities comprise the following type of attacks: \textit{ftp\_write, guess\_passwd, multihop, phf, imap, spy, warezclient, warezmaster, named, xsnoop, xlock, sendmail, worm, snmpgetattack, snmpguess}. Finally, the \textit{U2R (User to Root)} class comprises the following privilege escalation attacks: \textit{perl, rootkit, loadmodule, buffer\_overflow, httptunnel, ps, sqlattack, xterm}. Additionally, one can see in Table~\ref{tab:dist_clases_kdd} that NSL-KDD notably differs from the other networks datasets in terms of the attacks and normal traffic distributions, representing a more balanced problem than the previous network datasets, as opposite to what one would expect to observe in a real network environment.


\section{Reliable Network IDS (R-NIDS) methodology}
\label{sec:methods}


\begin{figure*}[!t]
\centering
\includegraphics[width=1\hsize,keepaspectratio=true]{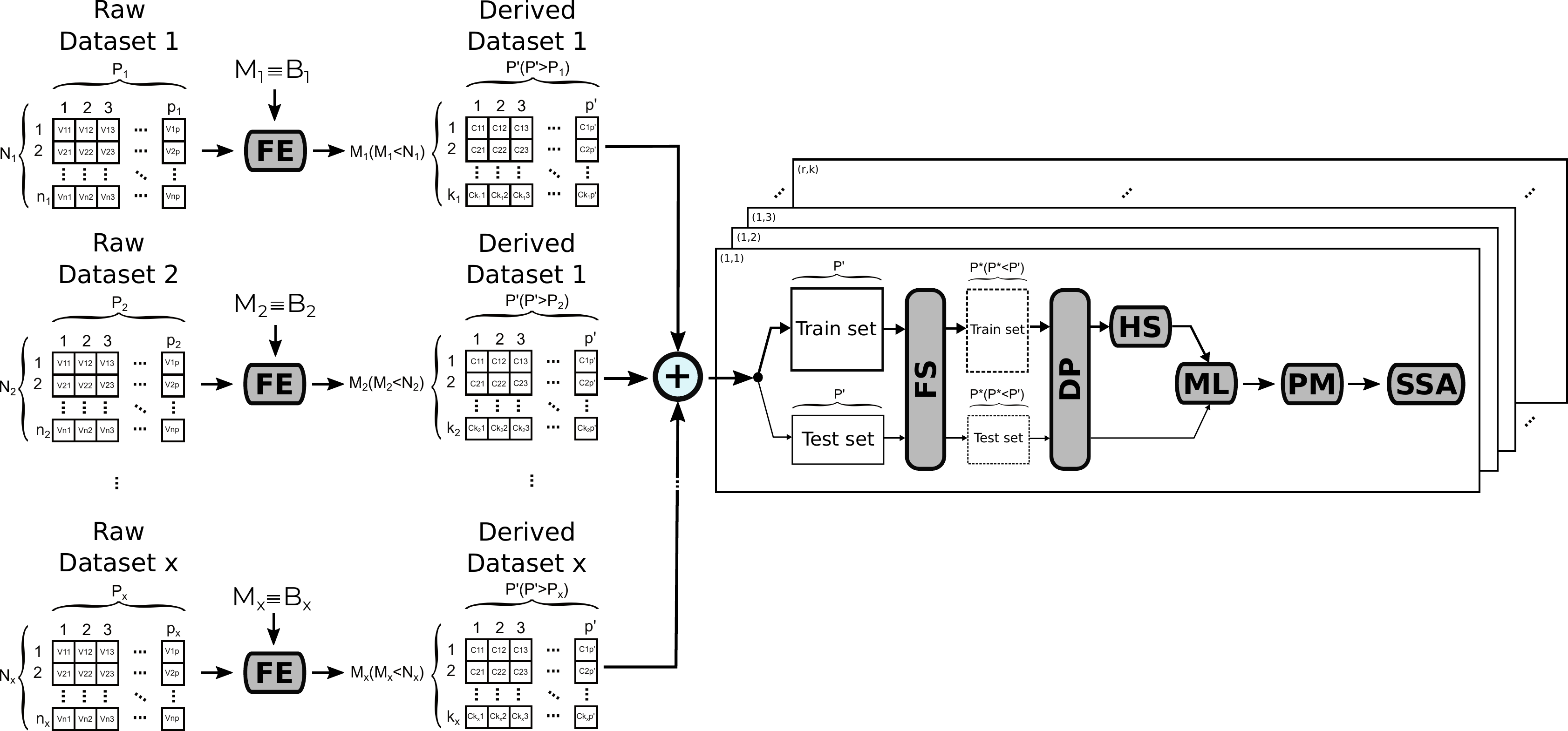}
\caption{R-NIDS framework and their main stages. FE: \textit{Feature Engineering}, FS: \textit{Feature Selection}, DP: \textit{Data Preprocessing}, HS: \textit{Hyperparameters Selecion}, ML: \textit{Machine Learning model}, PM: \textit{Performance Metric} and SSA: \textit{Statistical Significance Analysis}}.
\label{fig:ngl-nids}
\end{figure*}

Due to the continuous growing of security threats, more sophisticated methods and ways need to be devised to counteract against new samples of malicious security attacks, specially in the detection stage. Reliable-NIDS (R-NIDS) proposes a new methodology to achieve trustworthy and reliable NIDS solutions. On one hand, it provides a flexible and adaptable framework for a fair evaluation and comparison of NIDS solutions. On the other hand, and thanks to the new data integration approach proposed, makes it suitable to built enhanced ML models that perform correctly in network production environments. Modules and stages of R-NIDS are shown in Fig. \ref{fig:ngl-nids}.

On the right part of the figure, the main modules and stages to evaluate and compare NIDSs are shown. They are: the Feature Engineering (FE), the Feature Selection (FS), the Data Preprocessing (DP), the Hyper-parameter Selection (HP), the Machine Learning model (ML), the Performance Metric (PM) used and the Statistical Significante Analysis (SSA) performed. As it is emphasized in~\cite{Magan2020}, almost all the analyzed previous works either missed to include some of them or elude mentioning them, making it difficult to compare different NIDS solutions. 

\subsection{Batch-based FE for multipurpose data aggregation and integration: a row-wise approach}
\label{subsec:batchbased_fe}

As it can be seen in the left-hand side of Fig.~\ref{fig:ngl-nids}, R-NIDS is able to manage heterogeneous network datasets, to perform a data aggregation and to integrate them into only one overall NIDSs dataset. This is achieved thanks to the Feature as a Counter (FaaC) method~\cite{camacho_tackling_2014} applied in the FE module. FaaC converts the dataset into a derived dataset with completely new variables, which are counters on the values of the variables in the original dataset ($P_1, P_2,\cdots,P_x$ in the figure) for any given batch of raw network flows (e.g., a batch consisting of input flows withing each minute, or a batch consisting of a set of consecutive input flows). FaaC outputs data matrices of $M_x\times P^{'}$ dimension where the number of observations depends on the $M_x$ parameter but the number of derived variables, $P^{'}$, remains equals no matter the raw dataset used. $M_x$ will consequently determine the batch size, $B_x$, of raw network flows that are aggregated using the FaaC technique in order to obtain a derived dataset of $M_x$ observations. Thus, the higher the values of $M_x$, the smaller the size of the batch, and vice versa~\cite{magan-carrion_assessing_2022}. Furthermore, each feature $P^{'}$ in the derived dataset is normalized by the corresponding batch size used to process the raw dataset in order to guarantee that each new feature now lies in the $[0,1]$ range.

For a given classification task in supervised learning, the main goal is to accurately classify as many classes as the dataset has, which is the context of application in the current work. Concerning the dependent variable or output class in the derived dataset, the aggregation performed by the FaaC technique on a given batch of raw observations needs to account for specific situations, since the batch is likely to comprise more than one raw observations of different classes and, consequently, not making it trivial how to choose the class that the aggregated observation belongs to. In this work, we performed a simple procedure that consists of choosing the most predominant attack class counted from the raw observations within a batch. If no attack is present at all, then the aggregated observation is labeled as normal or background traffic. This solution penalizes those raw network datasets with balanced classes which, in fact, is not realistic since network traffic communications are imbalanced by nature (i.e., there should be much more normal or background traffic than attacks). It is the case of the NLS-KDD dataset. Instead, this procedure inherently balances positive (attacks) and negative (background) observations as we will see in the next section.  

All previous methods allow having a rectangular matrix per dataset where the variables' values are normalized counters in the $[0,1]$ range, making them easy to be managed as input by any ML model. Besides, this procedure provides the opportunity of customizing the way of integrating all the derived datasets into one joint dataset. The latter key functionality allows us to somehow mix different and heterogeneous network datasets to achieve reliable ML-based solutions. In this work, a simple row-wise data integration approach was employed as a case study, although any other way of combining different datasets could be set up thanks to the data aggregation and integration capabilities of the R-NIDS framework. In this way, one can think in to apply a column-wise approach or to smartly re-arrange and combine both, features and observations~\cite{magan-carrion_multivariate_2015}.




\subsection{Feature Selection}

Least Absolute Shrinkage and Selection Operator (LASSO) is used as a feature selection method in order to reduce the dimensionality of the datasets employed to train the classifiers, where only input features that the method considers relevant for the target outcome are retained. For this purpose, as a linear model the LASSO~\cite{Friedman2010} adds an $l_1$-penalty term to the minimization problem solved in a logistic regression model, as Equation~\ref{eq:lasso} defines:

\begin{equation}
\label{eq:lasso}
\bm{\hat{\beta}}_{\lambda} = \argmin_{\bm{\beta}} \enspace ||\bm{y}-f(\bm{\beta}X^T)||_2^2 + \lambda||\bm{\beta}||_1 \enspace ,
\end{equation}

\noindent where $\lambda$ is a hyper-parameter of the model which controls the strength of the regularization, i.e., the bigger the value of $\lambda$, the stronger the regularization and, consequently, less input variables would be retained, and vice versa.

\subsection{Data Pre-processing}
\label{sec:prepro}

In this work, a standard normalization procedure is applied to each numeric input variable, as depicted in Equation~\ref{eq:norm}, in such a way that normalized features have zero mean and unit variance:

\begin{equation}\label{eq:norm}
\bm{z_j} = \frac{\bm{x_j}-\mu_j}{\sigma_j}\;\;\;\;\;\; \forall{j\in[1,P]} \enspace ,
\end{equation}

\noindent where $\bm{x_j}$ is the raw input variable, $\mu_j$ and $\sigma_u$ are the mean and standard deviation of the given variable, and $\bm{z_j}$ is the standardized feature.

\subsection{Hyper-parameter Selection}
\label{subsec:hyper-parameter_selection}

Depending on the ML model used, the process of fitting a fairly optimized set of values for every model's hyper-parameters is not a trivial task. Traditionally, two well-known strategies have been used for this purpose: \emph{Grid Search}~\cite{divekar_benchmarking_2018,freeman_machine_2018} and \emph{Random Search}~\cite{Bergstra:2012}. 
Both approaches have some drawbacks in relation to computation efficiency and the limited exploration of the hyper-parameters search space. Therefore, they are not really suitable for real problems where accurate and efficient solutions are needed.

To overcome the limitations of the above strategies, the Bayesian hyper-parameter optimization~\cite{Snoek:2012} is used in this work. In this strategy, hyper-parameters search is modelled by an underlying Gaussian process in such a way that, on each step of the iterative search, the next hyper-parameters setting to be tested is the one with more uncertainty (higher variance) defined by this Gaussian process. Therefore, this strategy ensures that a pseudo-optimal hyper-parameters setting will be achieved in a few number iterations.

\subsection{Machine learning models}

Given a set of $n$ observations and $p$ variables describing each observation, in supervised ML, and more specifically in any classification task, the goal of a model is to learn in the best possible way the existing relationship between the dependent variable or output class $y^{(i)}$, and the independent variables $x_1^{(i)}$, ... , $x_p^{(i)}$, for any \emph{i}-th sample. In the context of this work, $y^{(i)}$ corresponds to the label or class associated to the \emph{i}-th observation, thus resulting in a multi-classification problem where $y^{(i)} \in [0,...,k]$ (i.e, background or normal traffic is mapped to 0, and each attack is mapped to consecutive numbers starting from 1).

Next, a brief description of the models used as classifiers in this work is provided:


\begin{itemize}
    \item Multinomial Logistic Regression (LR). It is the simplest possible linear model~\cite{bishop:2006}, in which labels are obtained from a linear combination of the input features and the parameters $\beta_0$, ..., $\beta_p$. In a binary classification task, this parameter vector is learned by solving the minimization problem depicted in Equation~\ref{eq:logreg}:
    


\begin{equation}
\label{eq:logreg}
\bm{\hat{\beta}} = \argmin_{\bm{\beta}} \enspace ||\bm{y}-f(\bm{\beta}X^T)||_2^2 \enspace ,
\end{equation}

\noindent where  $f(\bm{\beta}X^T)$ is the logistic or sigmoid function. In order to deal with the multi-classification setting, the well-known \emph{One-vs-All} approach is employed in such a way that $k+1$ LR models are fitted to data, each one focusing on a specific instance of the data that corresponds to a binary classification problem (e.g., one LR model to distinguish background or normal traffic from everything else, another one to distinguish a port scanning attack from the rest, etc.). Based on these models, the final output of the model is obtained by using the \emph{softmax} function depicted in Equation~\ref{eq:mullassoprob}.


\begin{equation}
\label{eq:mullassoprob}
Pr(y^{(i)}=k|X)=\frac{Pr(y^{(i)}=k|X)}{\sum_{z=0}^k{Pr(y^{(i)}=z|X)}}  \enspace ,
\end{equation}
    
    \item Random Forest (RF). It is an ensemble of multiple decision trees that are fitted to different instances of the data (i.e, a random sample of the observations and the predictors)~\cite{Breiman:2001}. Once a RF model is trained, it computes the class or label of any new unseen observation as the majority class among the ones provided by each individual decision tree.
    
\end{itemize}

\subsection{Performance Metrics}
\label{subsec:metrics}

Evaluating and comparing a ML-based NIDS solution relies on the use of performance metrics. However, the suitability of a metric depends on the application context. Thus, in this work we chose the AUC (Area Under the Curve) to evaluate ML-based NDISs when dealing with imbalanced data. Realistic network datasets are imbalanced by nature, having great differences among the number of the positive classes (attacks) in comparison with the negative one (normal traffic or background).

The AUC measures the area under the Receiver Operating Characteristic (ROC) \cite{salo_data_2018}, which draws the evolution of the TP (True Positive) rate versus the FP (False Positive) rate for different values of the classifying threshold. Thus, AUC=1 means that the solution is able to correctly classify all the observations while AUC $\sim$ 0.5 means that the classifier performs randomly.

Additionally, an AUC weighted average is also computed as follows: for each class $i=0,\ldots,k$, the $weighted\_avg(AUC_i)$ computes the weighted average of $AUC_i$ times $q_i$ (the number of true observations of each class), with $Q$ being the total number of observations. Equation~\ref{eq:wa} introduces the AUC weighted average. Note that this equation can be easily extended to other performance metrics. 

\begin{equation}
    \label{eq:wa}
    weighted\_avg(AUC_i) = \frac{\sum_{i=0}^{k}{AUC_{i} \times q_{i}}}{Q}
\end{equation}

\subsection{Significance analysis}\label{sect:statisticAlanalysis}

In this work, a statistical study on the obtained results was performed to assess statistical certainty on our conclusions, at 95\% confidence level. To correctly address it, we followed the recommendations given in~\cite{stapor2021} for the fair experimental evaluation of classifiers. In particular, we perform the Wilcoxon Signed-Rank Test~\cite{HWC13} in the pairwise comparisons of the accuracy of the two studied classifiers on all classes. Wilcoxon Signed-Rank Test is a non-parametric version of the paired T-test.

In all tables included in Section~\ref{sec:results} of this document, figures in {\bf bold font} stand for statistically significantly better results with respect to the compared classifier, at 95\% confidence level. 


\section{Experimental design}
\label{sec:experiments}
This sections describes the set-up of the R-NIDS framework introduced in Section~\ref{sec:methods} at the same time that provides a detailed description of the evaluation strategy employed to test the performance of the R-NIDS we propose. 

\subsection{R-NIDS set-up}	
\label{subsec:r-nids}
	



\begin{figure*}[!hbt]
\centering
	\subfloat[UGR'16 derived dataset.]{\includegraphics[width=3in]{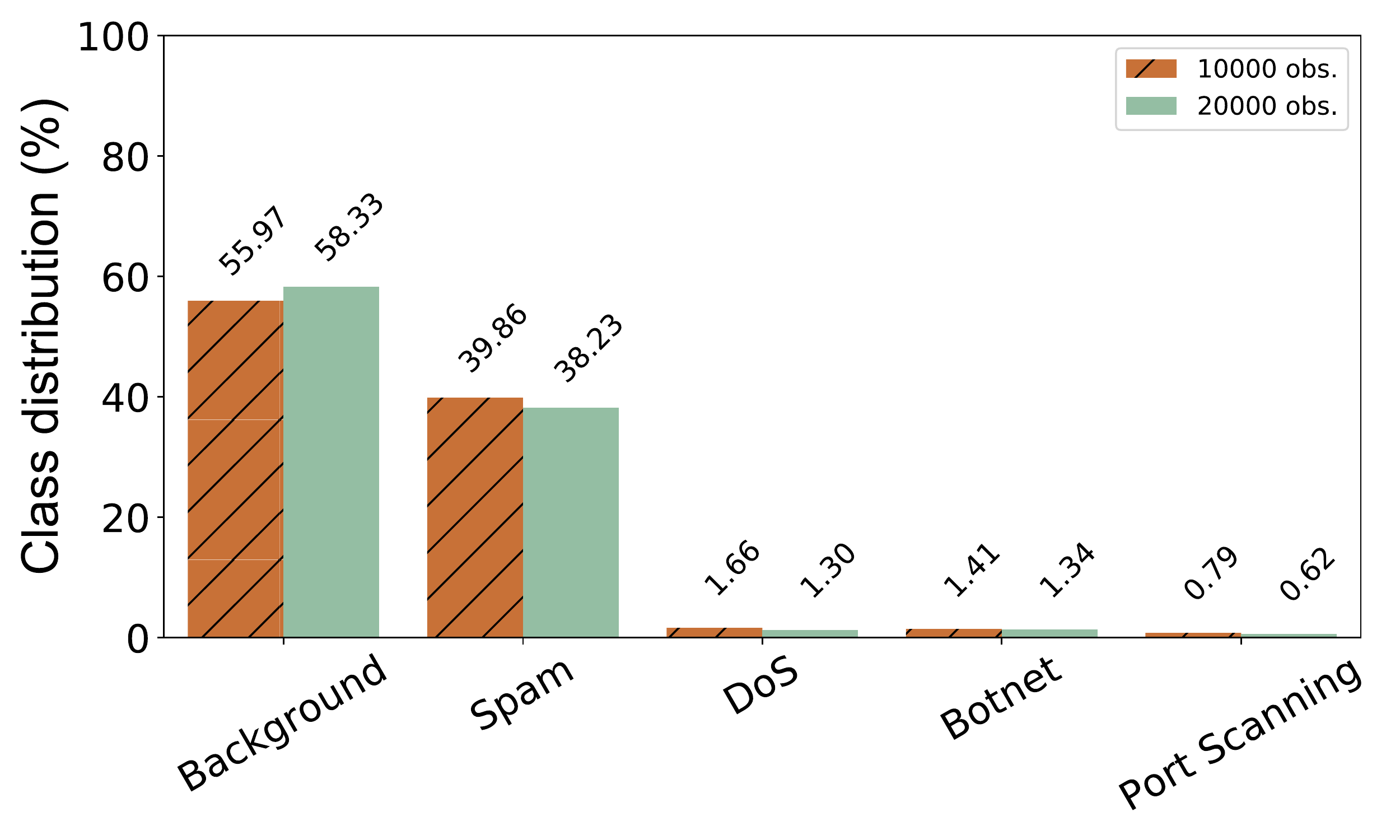}}
    \qquad
	\subfloat[UNSW-NB15 derived dataset.]{\includegraphics[width=3in]{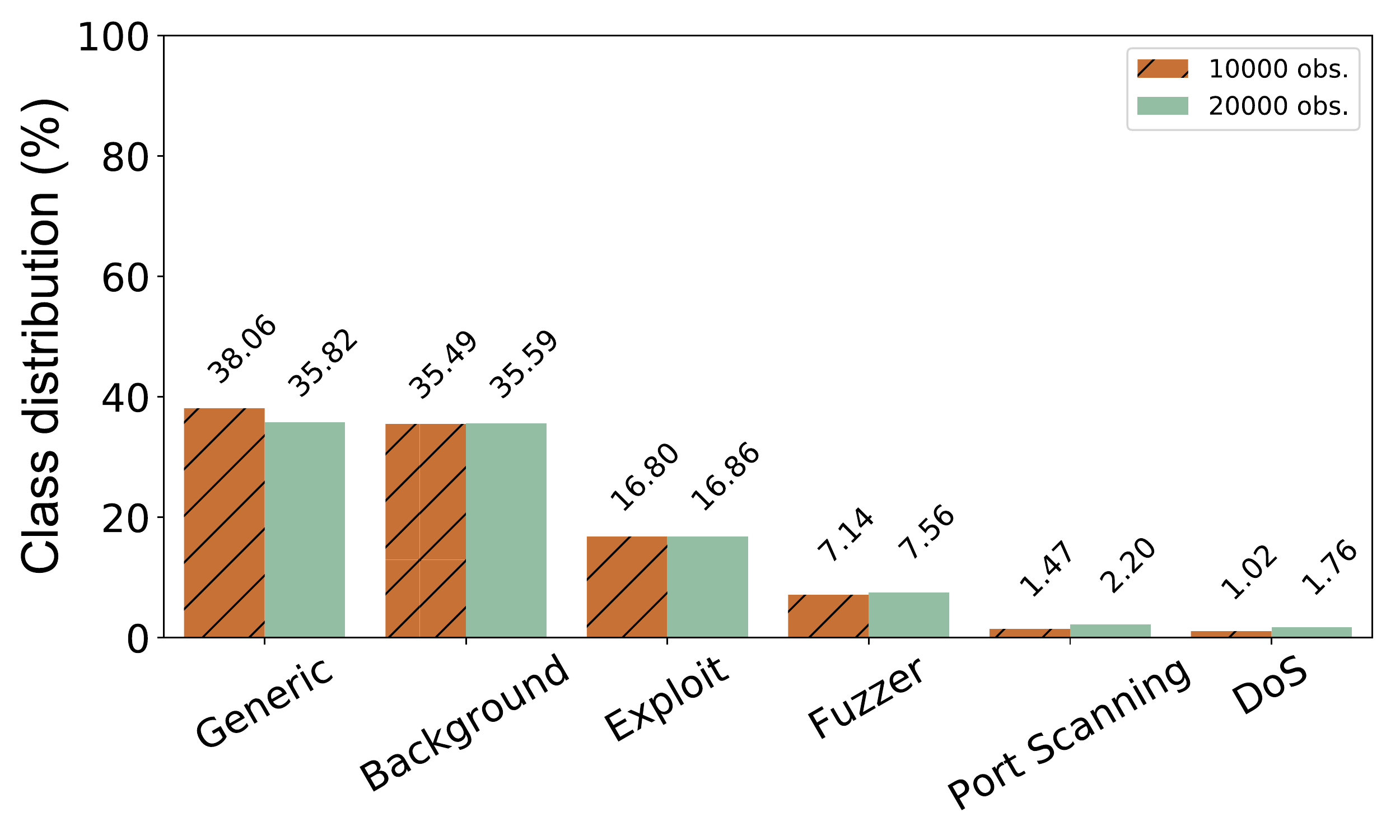}}
	\qquad
	\subfloat[NSL-KDD derived dataset.]{\includegraphics[width=3in]{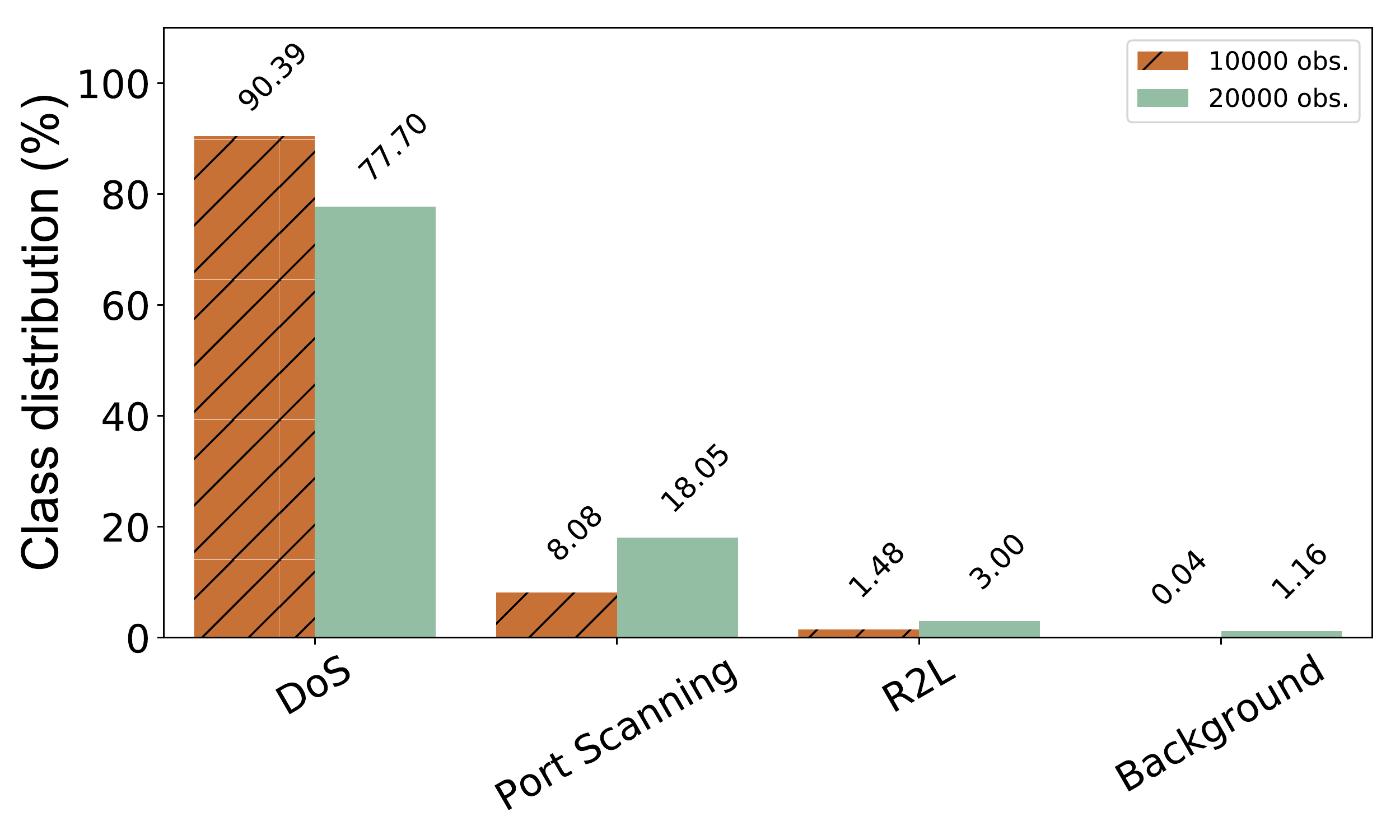}}
	\qquad
	\subfloat[UNK21 row-wise aggregated dataset.]{\includegraphics[width=3in]{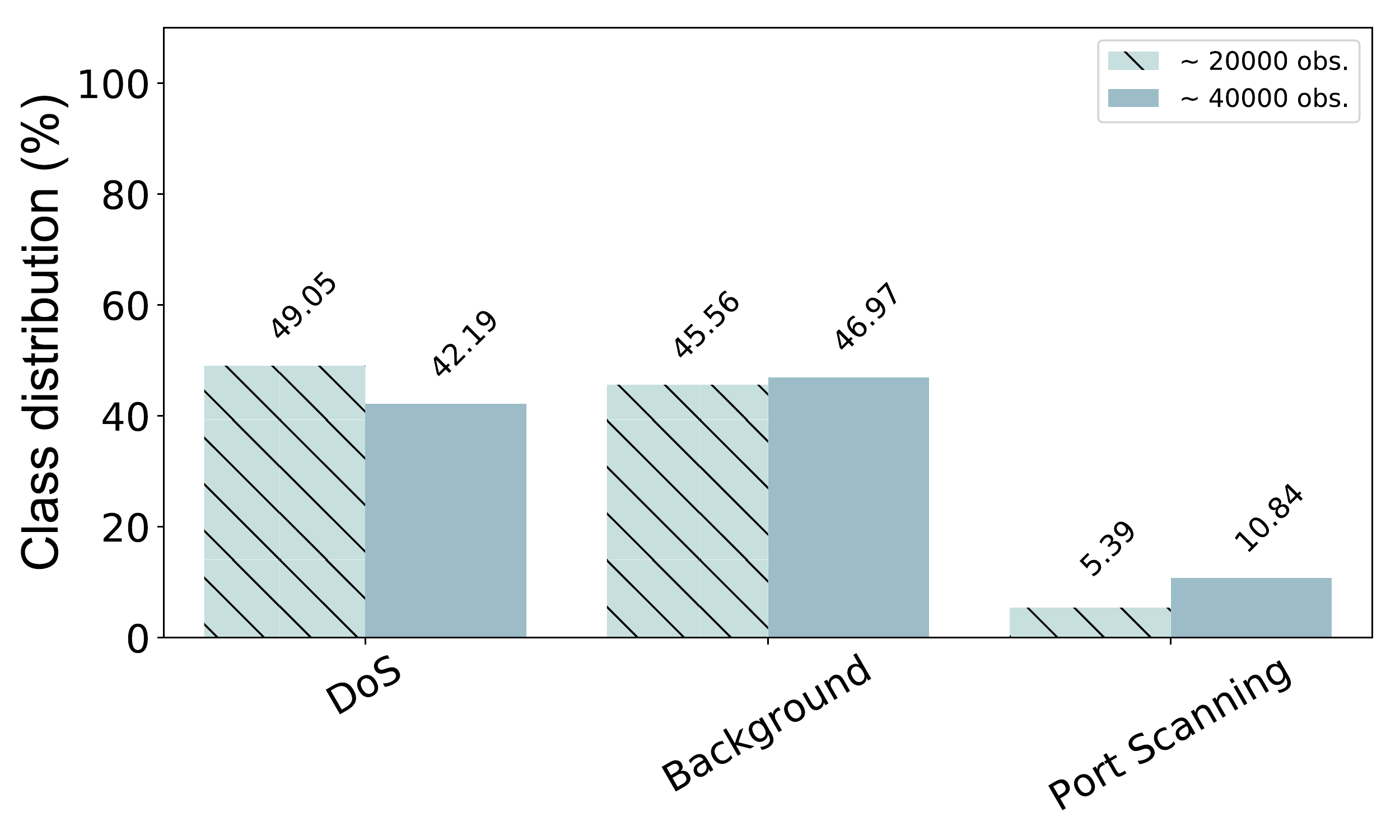}}
\caption{Class distribution on the derived datasets obtained by applying the FaaC method to the raw datasets included in Section~\ref{sec:datasets} ---see (a), (b) and (c)--- and on the UNK21 row-wise aggregated dataset proposed in this paper ---see (d)---.}
\label{fig:classes_derived_datasets}
\end{figure*}

As the proposed R-NIDS methodology establishes, we first apply the FaaC feature engineering method on the original network datasets to obtain the derived ones. The application of FaaC requires building batches of observations, wherein the counters on the values of the variables are applied. With the aim of making a fair comparison on the performance of the ML-based solutions built in this study, authors decided to go for two versions of the derived datasets in such a way that the generated number of observations ($M_1$, ... , $M_x$) is fixed: either 10,000 or 20,000 observations. Since the number of samples in the raw network datasets is different, the batch sizes ---($B_1$, ... , $B_x$)--- used to aggregate the raw observation into new ones, that are counters of these ones, differ. These batch sizes are summarized as follows:

\begin{itemize}
    \item 10,000 observations derived datasets: the batch size used to generate a derived dataset with this number of samples was \{395.082, 254, 14\} for the UGR'16, UNSW-NB15 and NSL-KDD raw datasets, respectively.
    \item 20,000 observations derived datasets: the batch size used to generate a derived dataset with this number of samples was \{197.541, 127, 7\} for the UGR'16, UNSW-NB15 and NSL-KDD raw datasets, respectively.
\end{itemize}

Furthermore, the application of the FaaC technique \mbox{using} the setting described on each network dataset will also modify the class distribution within each dataset (see Section~\ref{subsec:batchbased_fe}). In this sense, Figs.~\ref{fig:classes_derived_datasets}a-\ref{fig:classes_derived_datasets}c visualize the class distribution on both derived datasets for each of the three raw datasets. We can see how FaaC tends to balance them for the corresponding derived datasets of the UGR'16 and UNSW-NB15, that is, the percentages of all positive (attacks) classes and the negative (background) one are similar. However, we find the opposite behavior for the NSL-KDD dataset, mainly represented by observations of \emph{DoS} attack and almost no background or normal traffic. This may be an indication which supports the procedure used to create this dataset using synthetic, i.e., unrealistic network flows.


Next, the R-NIDS framework aims at creating one single derived dataset that combines all the derived datasets generated by using the FaaC technique. For this purpose, only observations that correspond to \emph{Background}, \emph{DoS} or \emph{Port Scanning} classes (i.e., the ones in common among the three datasets) are considered during the aggregation process. Then, and thanks to the variables homogenization achieved with the application of the FaaC technique, a simple row-wise aggregation procedure is used to merge all the observations of these three derived datasets, which met the criteria mentioned before, onto a single new integrated dataset that is hereafter called UNK21, taking its name from the capital letters of the three involved network datasets. Two versions of this dataset are obtained depending on the number of observations that the derived datasets have. Thus, a UNK21 $\sim$20,000 observations dataset was built by following a row-wise data aggregation of the 10,000 observations derived datasets. Similarly, for the UNK21 $\sim$40,000 observations. It is worth remembering at this point that only classes that are present in the three datasets are considered to generate UNK21. 

The class distribution within the UNK21 dataset is shown in Fig.~\ref{fig:classes_derived_datasets}d. There is a balanced representation of the \emph{Background} and \emph{DoS} classes, although the \emph{Port scanning} class is still underrepresented. Finally, a last and key normalization step is performed in such a way that each variable of any observation within the UNK21 dataset is normalized by the batch size used to generate it according to the original raw network dataset. This procedure is performed as shown in Equation~\ref{eq:observ_norm}:

\begin{equation}
\label{eq:observ_norm}
norm(\bm{x}) = \begin{cases}
\frac{\bm{x}}{B_{UGR'16}}, & \text{if }\bm{x} \in \text{UGR'16 dataset}\\
\frac{\bm{x}}{B_{UNSW-NB15}}, & \text{if }\bm{x} \in \text{UNSW-NB15 dataset}\\
\frac{\bm{x}}{B_{NSL-KDD}}, & \text{if }\bm{x} \in \text{NSK-KDD dataset}\\
\end{cases}
\end{equation}

\noindent
where $\bm{x}=x_1,\cdots, x_p$ corresponds with the derived variables (counters) to be normalized and the $B_{dataset}$ is the batch size used to normalize them. 

The presented normalization step guarantees that variables' values of all the observations within the UNK21 dataset are normalized counters in the $[0,1]$ range, regardless their dataset of origin. Consequently, the ML-based solution build on this UNK21 dataset could be a more robust and reliable NIDS than others using the classic one-single-dataset approach. Moreover, this system will be ready to use in production and for generic network environments, since the application of this normalization procedure ensures that variables within new unseen observations will also lie in the same range as the expected by the trained models (i.e., in the $[0,1]$ range).


\subsection{Evaluation strategy}
In this study, two evaluation settings were considered:

\begin{itemize}
    \item Analysis of one single dataset: in this setting where ML-based solutions for NIDSs are built and tested on a single network dataset, the entire analysis was carried out using a 5-fold cross-validation strategy ($k=5$ in Fig.\ref{fig:ngl-nids}), where the complete dataset is divided into 5 folds of equal sizes, assuring that the class distribution remains the same within each fold. This strategy trains the ML models  proposed as classifiers (which includes a feature selection through LASSO and a z-score data normalization steps before fitting the model to the data) in 4 out of the 5 folds and finally evaluates its performance on the test fold (the fifth one) left apart. Then, this procedure is repeated iteratively by rotating the folds used to train and test the classifiers. Furthermore, the fold partitioning process is repeated 20 times ($r=20$ in Fig. \ref{fig:ngl-nids}) in order to guarantee the randomness of this process.
    
    \item Analysis across datasets: the analysis carried out in this setting is much more simple, since it aims at training the ML models proposed as classifiers in one out of the three network datasets considered, and evaluate its performance in the remaining two datasets, not used to train the models. This procedure is repeated iteratively by rotating the complete dataset used to train the classifiers. 
\end{itemize}
 
 The former setting aims at performing the classical ML-based NIDS evaluation that is typically found in the literature, where the community focuses on developing ML-based solutions using one single network dataset, thus forgetting to double-check how well the ML-based solution generalizes with different datasets (i.e., gathered in a different network environments). The latter setting tries to complete the big picture by showing the performance of these kind of ML-based solutions when these models are used to predict attacks on a different network environment than the one used to train them. In both of these settings, the hyper-parameters of the ML models considered in this study were learned through Bayesian optimization in the training set, whose benefits were introduced in section~\ref{subsec:hyper-parameter_selection}. Finally, the AUC previously introduced in section~\ref{subsec:metrics} is chosen as a performance metric because it is found suitable when dealing with imbalanced network datasets for comparing and evaluating NIDSs.

\section{Results and discussion}
\label{sec:results}

This section summarizes the results of the analysis and discusses the main findings. We first compare the performance of LR and RF classifiers when they are individually built on one of the above introduced network datasets: UGR'16, USNW-NB15, and NSL-KDD. Tables~\ref{tab:results_cvugr16}-\ref{tab:results_cvkdd} show the AUC performance obtained by these ML models on each network  dataset, individually, using the 10,000 or the 20,000 observations derived datasets (as mentioned in Section~\ref{sect:statisticAlanalysis}, significant differences are emphasized in \textbf{bold font}). This is the most common approach found in the literature and, as expected, ML-based solutions tend to achieve high accuracy rates. If we focus on the weighted average AUC ---which takes into account how accurate the model is on each single class together with the number of observations per class\mbox{---,} on average ML-based solutions achieve an AUC $\approx$ 0.96 and AUC $\approx$ 0.98 in the UGR'16 and UNSW-NB15 datasets, respectively (highly accurate performance). However, in the NSL-KDD dataset the best weighted average AUC is approximately 0.76, which suggests that ML models struggle to learn correctly the attack patterns of this network dataset and could be an indicator of this dataset not having enough quality in the synthetic network flows. Concerning the two ML models tested, the non-linear model (RF) clearly perfoms better than the linear one (LR) in the UGR'16 and UNSW-NB15 datasets, with significant differences in almost all comparisons (in 24 out of 26 cases). However, the LR outperforms RF in the case of NSL-KDD dataset, specially when 10,000 observations are used.

\begin{table}[!t]
    \centering
    \caption{Average AUC performance obtained in the UGR'16 dataset using a 20 repetitions of 5-fold cross-validation evaluation strategy.}
    \label{tab:results_cvugr16}
\begin{tabular}{c|c|c|cccccc}
\toprule
\textbf{Dataset} & \textbf{Model} & \textbf{Class} &    \textbf{AUC}  \\ 
\midrule
\multirow{12}{*}{\rotatebox[origin=c]{90}{\textbf{UGR'16} 10,000 obs.}}
&\multirow{6}{*}{LR}          & Background             & 0.894  \\
&                             & Botnet                 & 0.986  \\
&                             & DoS                    & 0.977  \\
&                             & Port Scanning          & 0.980  \\
&                             & Spam                   & 0.893  \\
&                             & \cellcolor[gray]{0.8}\textit{Weighted avg.} & \cellcolor[gray]{0.8}\textit{0.897}\\ 
\cline{2-4}
&\multirow{6}{*}{RF}          & Background             & {\bf 0.962}  \\
&                             & Botnet                 & {\bf 0.998} \\
&                             & DoS                    & {\bf 0.987}  \\
&                             & Port Scanning          & {\bf 0.997}  \\
&                             & Spam                   & {\bf 0.964}  \\
&                             & \cellcolor[gray]{0.8}\textit{Weighted avg.} & \cellcolor[gray]{0.8}\textit{{\bf 0.964}} \\ 
\midrule
\multirow{12}{*}{\rotatebox[origin=c]{90}{\textbf{UGR'16} 20,000 obs.}}
&\multirow{6}{*}{LR}          & Background             & 0.893  \\
&                             & Botnet                 & 0.993  \\
&                             & DoS                    & 0.985  \\
&                             & Port Scanning          & 0.988  \\
&                             & Spam                   & 0.892  \\
&                             & \cellcolor[gray]{0.8}\textit{Weighted avg.} & \cellcolor[gray]{0.8}\textit{0.896} \\ 
\cline{2-4}
&\multirow{6}{*}{RF}          & Background             & {\bf 0.962}  \\
&                             & Botnet                 & {\bf 0.998}  \\
&                             & DoS                    & 0.987  \\
&                             & Port Scanning          & {\bf 0.997}  \\
&                             & Spam                   & {\bf 0.963}  \\
&                             & \cellcolor[gray]{0.8}\textit{Weighted avg.} & \cellcolor[gray]{0.8}{\textit{{\bf 0.964}}} \\ 
\bottomrule
\end{tabular}
\end{table}

\begin{table}[!t]
    \centering
    \caption{Average AUC performance obtained in the UNSW-NB15 dataset using a 20 repetitions of 5-fold cross-validation evaluation strategy.}
    \label{tab:results_cvnb15}
\begin{tabular}{c|c|c|cc}
\toprule
\textbf{Dataset} & \textbf{Model} & \textbf{Class} & \textbf{AUC} \\
\midrule
\multirow{14}{*}{\rotatebox[origin=c]{90}{\textbf{USNW-NB15} 10,000 obs.}}
&\multirow{7}{*}{LR}          & Background             & 0.994   \\
&                             & DoS                    & 0.954   \\
&                             & Exploit                & 0.937   \\
&                             & Fuzzer                 & 0.962   \\
&                             & Generic                & 0.999   \\
&                             & Port Scanning          & 0.875  \\
&                             & \cellcolor[gray]{0.8}\textit{Weighted avg.} & \cellcolor[gray]{0.8}\textit{0.982} \\ 
\cline{2-4}
&\multirow{7}{*}{RF}          & Background             & {\bf 0.998}   \\
&                             & DoS                    & {\bf 0.978}   \\
&                             & Exploit                & {\bf 0.961}  \\
&                             & Fuzzer                 & {\bf 0.965}  \\
&                             & Generic                & {\bf 0.999}   \\
&                             & Port Scanning          & {\bf 0.930}   \\
&                             & \cellcolor[gray]{0.8}\textit{Weighted avg.} & \cellcolor[gray]{0.8}\textit{{\bf 0.988}}  \\ 
\midrule
\multirow{14}{*}{\rotatebox[origin=c]{90}{\textbf{USNW-NB15} 20,000 obs.}}
&\multirow{7}{*}{LR}          & Background             & 0.985 &  \\
&                             & DoS                    & 0.885 &  \\
&                             & Exploit                & 0.904 &  \\
&                             & Fuzzer                 & {\bf 0.956} &  \\
&                             & Generic                & 0.994 &  \\
&                             & Port Scanning          & 0.882 &  \\
&                             & \cellcolor[gray]{0.8}\textit{Weighted avg.} & \cellcolor[gray]{0.8}\textit{0.968} \\ 
\cline{2-4}
&\multirow{7}{*}{RF}          & Background             & {\bf 0.992} &  \\
&                             & DoS                    & {\bf 0.920} &  \\
&                             & Exploit                & {\bf 0.933} &  \\
&                             & Fuzzer                 & 0.953 &  \\
&                             & Generic                & {\bf 0.996} &  \\
&                             & Port Scanning          & {\bf 0.911} &  \\
&                             & \cellcolor[gray]{0.8}\textit{Weighted avg.} & \cellcolor[gray]{0.8}{\textit{{\bf 0.977}}}  \\ 
\bottomrule
\end{tabular}
\end{table}

\begin{table}[!t]
    \centering
    \caption{Average AUC performance obtained in the NSL-KDD dataset using a 20 repetitions of 5-fold cross-validation evaluation strategy.}
    \label{tab:results_cvkdd}
\begin{tabular}{c|c|c|cccccc}
\toprule
\textbf{Dataset} & \textbf{Model} & \textbf{Class} & \textbf{AUC} \\
 
\midrule
\multirow{10}{*}{\rotatebox[origin=c]{90}{\textbf{NSL-KDD} 10,000 obs.}}
&\multirow{5}{*}{LR}          & Background             & {\bf 0.983}   \\
&                             & DoS                    & {\bf 0.761}   \\
&                             & Port Scanning          & {\bf 0.731}  \\
&                             & R2L                    & {\bf 0.961}  \\
&                             & \cellcolor[gray]{0.8}\textit{Weighted avg.} & \cellcolor[gray]{0.8}\textit{{\bf 0.761}} \\ 
\cline{2-4}
&\multirow{5}{*}{RF}          & Background             & 0.950   \\
&                             & DoS                    & 0.723   \\
&                             & Port Scanning          & 0.698   \\
&                             & R2L                    & 0.940   \\
&                             & \cellcolor[gray]{0.8}\textit{Weighted avg.} & \cellcolor[gray]{0.8}\textit{0.724} \\ 
\midrule
\multirow{10}{*}{\rotatebox[origin=c]{90}{\textbf{NSL-KDD} 20,000 obs.}}
&\multirow{6}{*}{LR}          & Background             & {\bf 0.985}   \\
&                             & DoS                    & {\bf 0.749}   \\
&                             & Port Scanning          & 0.710   \\
&                             & R2L                    & 0.916   \\
&                             & \cellcolor[gray]{0.8}\textit{Weighted avg.} & \cellcolor[gray]{0.8}\textit{{\bf 0.750}} \\ 
\cline{2-4}
&\multirow{6}{*}{RF}          & Background            & 0.979   \\
&                             & DoS                    & 0.738   \\
&                             & Port Scanning          & 0.707  \\
&                             & R2L                    & {\bf 0.920}   \\
&                             & \cellcolor[gray]{0.8}\textit{Weighted avg.} & \cellcolor[gray]{0.8}{\textit{0.741}} \\ 
\bottomrule
\end{tabular}
\end{table}

\begin{table*}[!t]
    \centering
    \renewcommand{\arraystretch}{1.4}
    \caption{AUC performance obtained for the setting where a given model (LR/RF) is trained using a complete dataset and fully evaluated on a different and independent dataset. Cells highlighted with a gray background colour correspond to the results obtained for each dataset-specific cross-validation setting shown in Tables \ref{tab:results_cvugr16}-\ref{tab:results_cvkdd}.} 
    \label{tab:results_traintest_all}
\resizebox{\textwidth}{!}{%
\begin{tabular}{c|c|c|ccc|ccc|ccc}
\toprule
        \textbf{Size}   & \textbf{Model}   & \diagbox{\textbf{Train data}}{\textbf{Test data}}     
                    & \multicolumn{3}{c|}{\textbf{UGR'16}} & \multicolumn{3}{c|}{\textbf{USNW-NB15}} & \multicolumn{3}{c}{\textbf{NLS-KDD}}                                                    \\ \cline{4-12}
    &   & \multirow{-2}{*}{\textbf{}} & \textbf{Background}         & \textbf{DoS}                & \textbf{Port Scanning}               
                            & \textbf{Background}         & \textbf{DoS}                & \textbf{Port Scanning}                                                     
                            & \textbf{Background}         & \textbf{DoS}                & \textbf{Port Scanning}               \\ \bottomrule
    
    \toprule
    \multirow{6}{*}{\rotatebox[origin=c]{90}{10,000 obs.}}    & \multirow{3}{*}{LR}  
    & \textbf{UGR'16}             & \cellcolor[gray]{0.8}0.894 & \cellcolor[gray]{0.8}0.977 & \cellcolor[gray]{0.8}0.980 & 0.490 & 0.500 & 0.494 & 0.500 & 0.500 & 0.500  \\ \cline{3-12}
    &   & \textbf{USNW-NB15}          & 0.342 & 0.499 & 0.780 & \cellcolor[gray]{0.8}0.994 & \cellcolor[gray]{0.8}0.954 & \cellcolor[gray]{0.8}0.875 & 0.712 & 0.489 & 0.434  \\ \cline{3-12}
    &   & \textbf{NLS-KDD}            & -     & 0.500 & 0.282 & -               & 0.869 & 0.473 & \cellcolor[gray]{0.8}0.983 & \cellcolor[gray]{0.8}0.761 & \cellcolor[gray]{0.8}0.731  \\ \cline{2-12}
    
    & \multirow{3}{*}{RF}   
    & \textbf{UGR'16}             & \cellcolor[gray]{0.8}0.962 & \cellcolor[gray]{0.8}0.987 & \cellcolor[gray]{0.8}0.997 & 0.695 & 0.731 & 0.565 & 0.663 & 0.629 & 0.474  \\ \cline{3-12}
    &   & \textbf{USNW-NB15}          & 0.508 & 0.832 & 0.499 & \cellcolor[gray]{0.8}0.998 & \cellcolor[gray]{0.8}0.978 & \cellcolor[gray]{0.8}0.930 & 0.894 & 0.571 & 0.543  \\ \cline{3-12}
    &   & \textbf{NLS-KDD}            & - & 0.586 & 0.150 & - & 0.942 & 0.157 & \cellcolor[gray]{0.8}0.950 & \cellcolor[gray]{0.8}0.723 & \cellcolor[gray]{0.8}0.698  \\ \hline \hline
    
    \multirow{6}{*}{\rotatebox[origin=c]{90}{20,000 obs.}} & \multirow{3}{*}{LR}  
    & \textbf{UGR'16}             & \cellcolor[gray]{0.8}0.893 & \cellcolor[gray]{0.8}0.985 & \cellcolor[gray]{0.8}0.988 & 0.497 & 0.500 & 0.499 & 0.478 & 0.500 & 0.500  \\ \cline{3-12}
    &   & \textbf{USNW-NB15}          & 0.293 & 0.502 & 0.870 & \cellcolor[gray]{0.8}0.985 & \cellcolor[gray]{0.8}0.885 & \cellcolor[gray]{0.8}0.882 & 0.623 & 0.595 & 0.427  \\ \cline{3-12}
    &   & \textbf{NLS-KDD}            & 0.747 & 0.689 & 0.325 & 0.623 & 0.740 & 0.524 & \cellcolor[gray]{0.8}0.985 & \cellcolor[gray]{0.8}0.749 & \cellcolor[gray]{0.8}0.710  \\ \cline{2-12}
    
    & \multirow{3}{*}{RF}  
    & \textbf{UGR'16}             & \cellcolor[gray]{0.8}0.962 & \cellcolor[gray]{0.8}0.987 & \cellcolor[gray]{0.8}0.997 & 0.506 & 0.392 & 0.536 & 0.329 & 0.460 & 0.562  \\ \cline{3-12}
    &   & \textbf{USNW-NB15}          & 0.630 & 0.657 & 0.120 & \cellcolor[gray]{0.8}0.992 & \cellcolor[gray]{0.8}0.920 & \cellcolor[gray]{0.8}0.911 & 0.756 & 0.554 & 0.549  \\ \cline{3-12}
    &   & \textbf{NLS-KDD}            & 0.579 & 0.454 & 0.398 & 0.827 & 0.460 & 0.562 & \cellcolor[gray]{0.8}0.979 & \cellcolor[gray]{0.8}0.738 & \cellcolor[gray]{0.8}0.707  \\ \bottomrule
\end{tabular}
}
\end{table*}

\begin{figure*}[!t]
\centering
	\subfloat[Background traffic.]{\includegraphics[width=8cm]{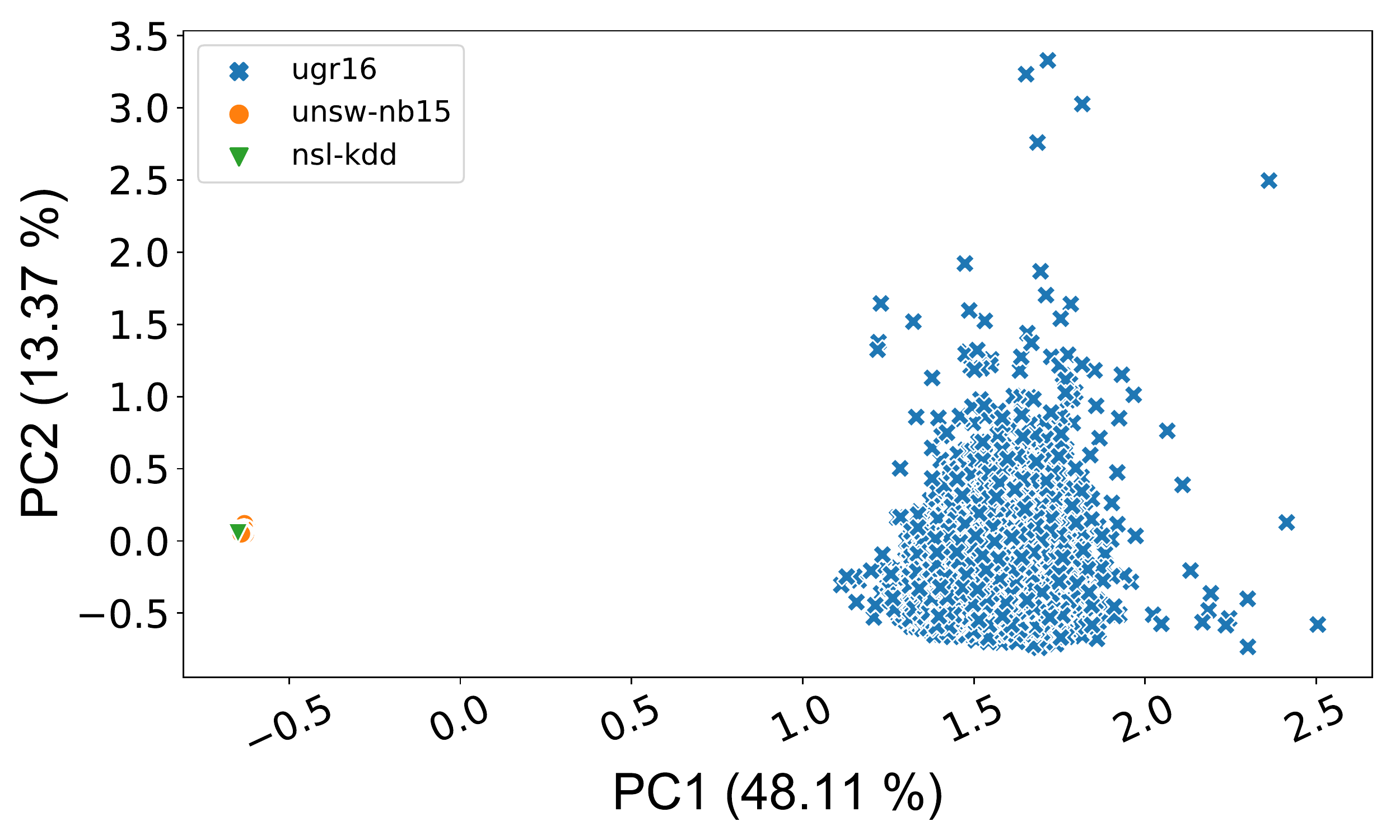}}
    \qquad
	\subfloat[Background traffic (details).]{\includegraphics[width=8cm]{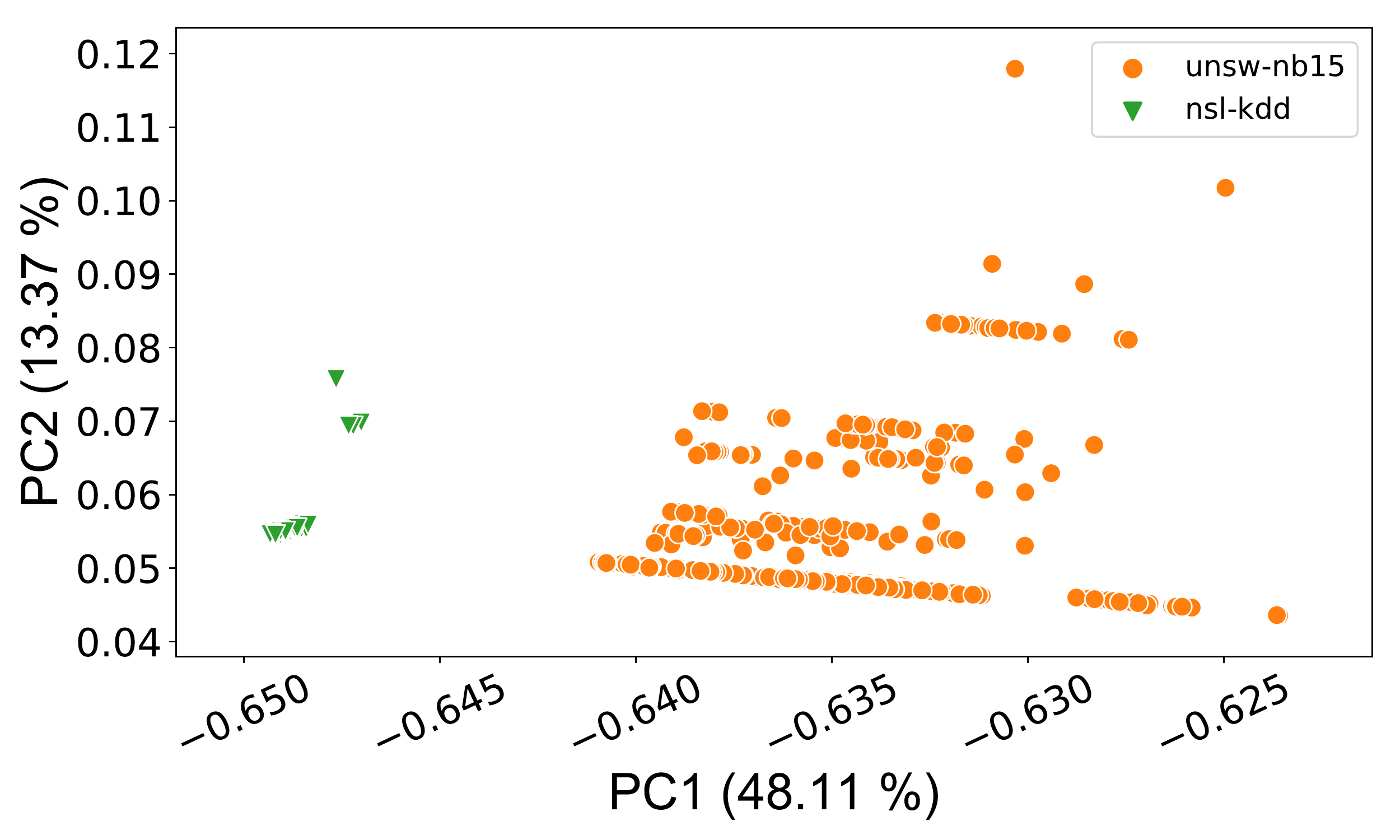}}
	\qquad
	\subfloat[Port Scanning attack.]{\includegraphics[width=8cm]{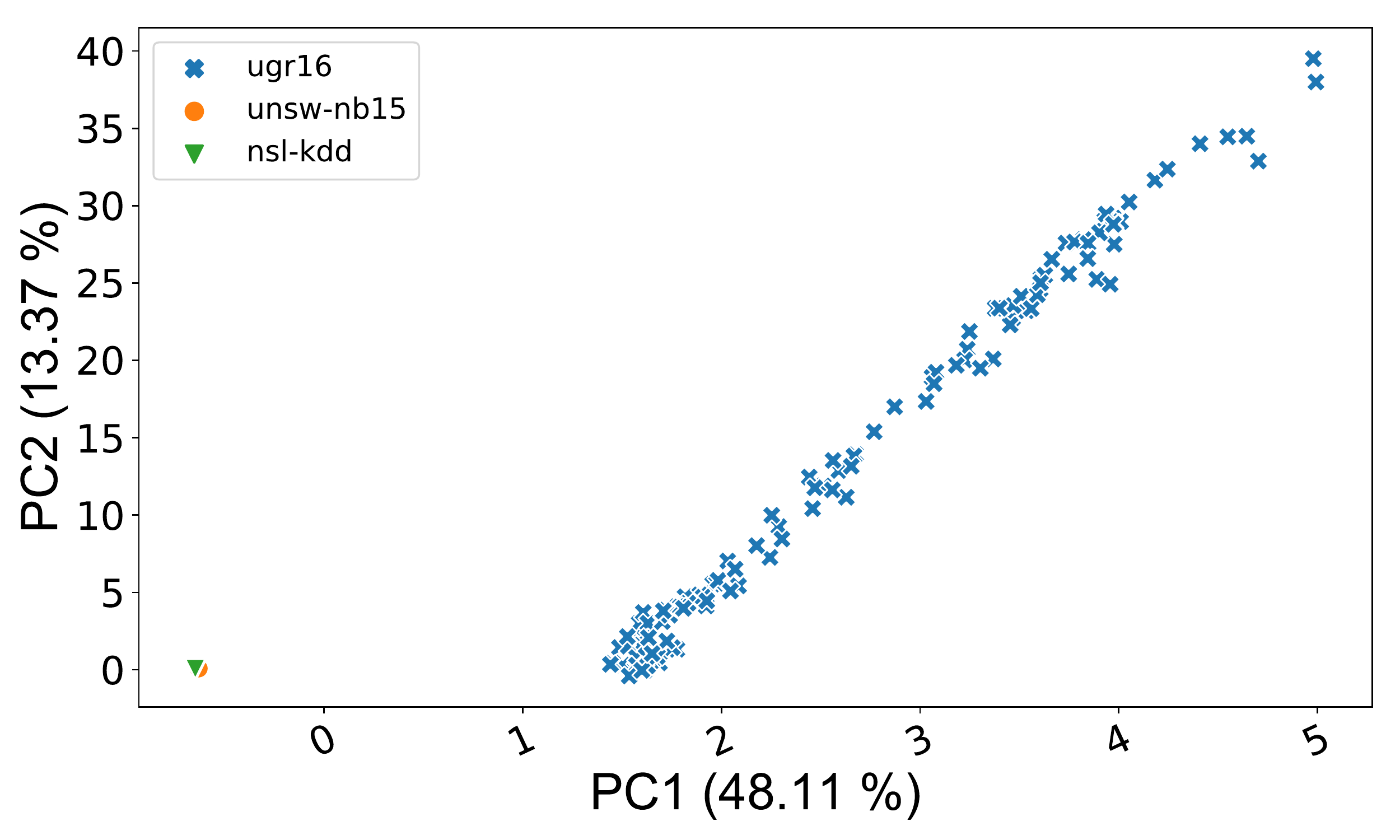}}
	\qquad
	\subfloat[Port Scanning attack (details).]{\includegraphics[width=8cm]{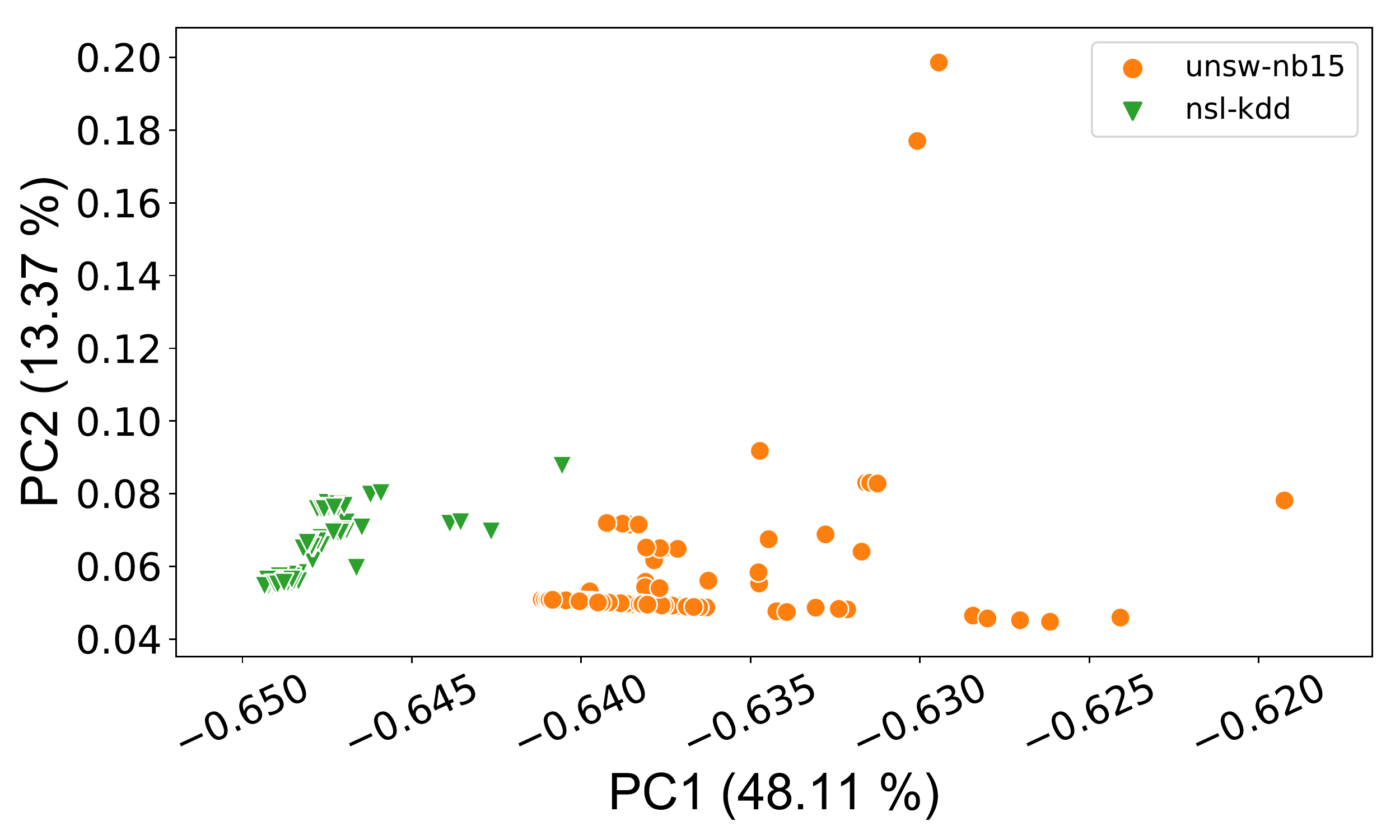}}
	\qquad
	\subfloat[DoS attack.]{\includegraphics[width=8cm]{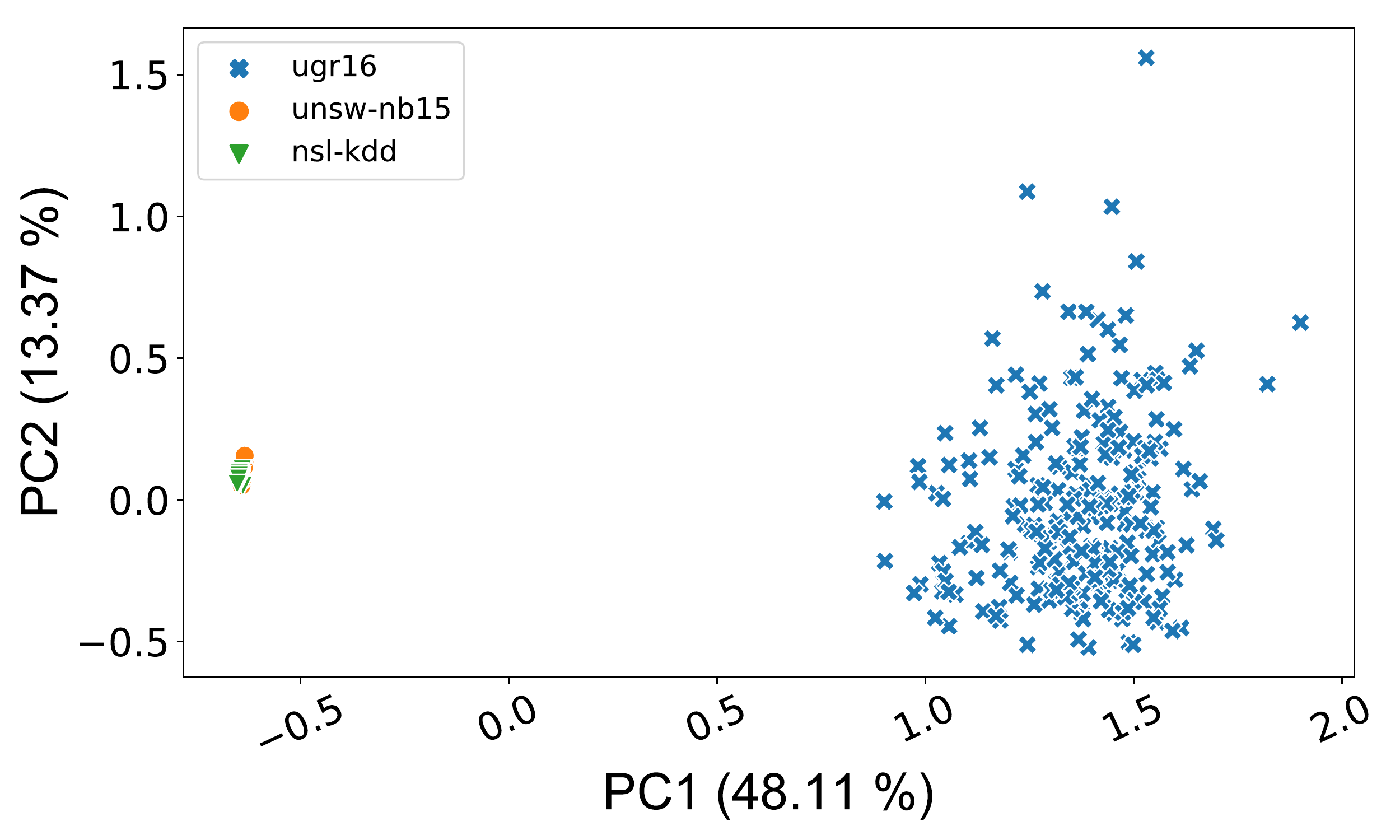}}
	\qquad
	\subfloat[DoS attack (details).]{\includegraphics[width=8cm]{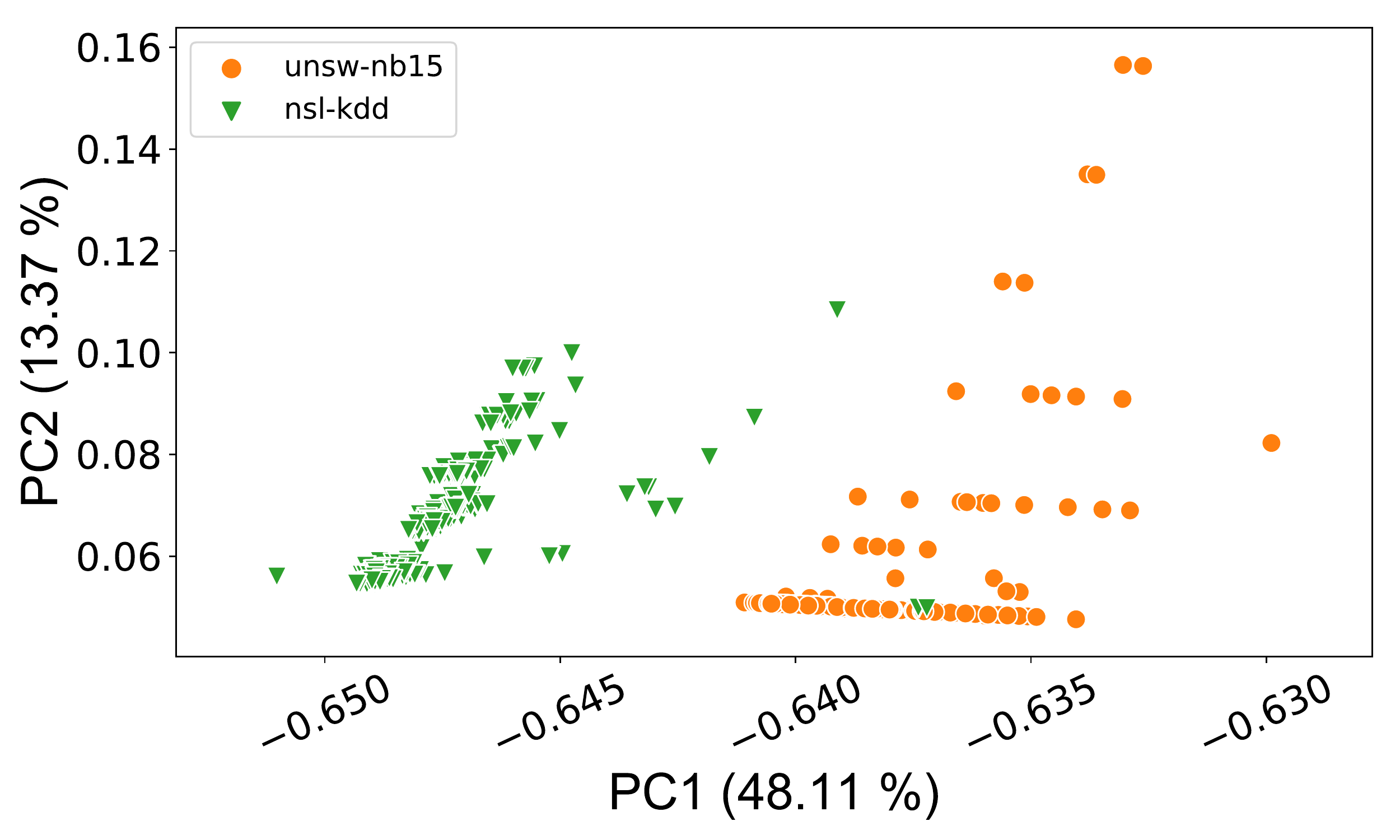}}
\caption{Results of the PCA analysis of the three NIDS datasets (UGR'16, UNSW-NB15, NSL-KDD), for each class.}
\label{fig:pca_analysis}
\end{figure*}

\begin{table*}[!t]
    \centering
    \renewcommand{\arraystretch}{1.4}
    \caption{Average AUC performance obtained in the UNK21 dataset using a 20 repetitions of 5-fold cross-validation evaluation strategy.}
    \label{tab:results_cvmix_zoom}
\begin{tabular}{c|c|c|c|c|c|c}
\toprule 
\textbf{Dataset} & \textbf{Model} & \textbf{Class} & \textbf{Overall} & \textbf{Only UGR'16} & \textbf{Only UNSW-NB15} & \textbf{Only NSL-KDD}\\
\hline

\multicolumn{1}{c|}{\multirow{8}{*}{\rotatebox[origin=c]{90}{\textbf{UNK21} $\sim$20,000 obs.}}}  
& \multicolumn{1}{c|}{\multirow{4}{*}{LR}}      
& \multicolumn{1}{c|}{Background} & 0.997 & 0.974 & 0.822  & {\bf 0.780}  \\
\multicolumn{1}{c|}{} & \multicolumn{1}{c|}{}                         
& \multicolumn{1}{c|}{DoS}        & {\bf 0.962} & 0.960 & 0.901  & {\bf 0.576}  \\
\multicolumn{1}{c|}{} & \multicolumn{1}{c|}{}  
& \multicolumn{1}{c|}{Port Scanning}        & 0.766 & 0.985 & 0.756  & {\bf 0.575}  \\ 
\multicolumn{1}{c|}{} & \multicolumn{1}{c|}{}  
& \multicolumn{1}{c|}{\cellcolor[gray]{0.8}\textit{Weighted avg.}} & \cellcolor[gray]{0.8}\textit{0.967} & \cellcolor[gray]{0.8}\textit{0.974} 
& \cellcolor[gray]{0.8}\textit{0.822}  & \cellcolor[gray]{0.8}\textit{{\bf 0.574}} \\ \cline{2-7} 
\multicolumn{1}{c|}{} 
& \multicolumn{1}{c|}{\multirow{4}{*}{RF}}      
& \multicolumn{1}{c|}{Background} & {\bf 0.998} & {\bf 0.983} & {\bf 0.892}  & 0.676  \\
\multicolumn{1}{c|}{} & \multicolumn{1}{c|}{}                         
& \multicolumn{1}{c|}{DoS}        & 0.960 & {\bf 0.978} & {\bf 0.937}  & 0.542  \\
\multicolumn{1}{c|}{} & \multicolumn{1}{c|}{}  
& \multicolumn{1}{c|}{Port Scanning}        & 0.765 & {\bf 0.995} & {\bf 0.849}  & 0.539 \\ 
\multicolumn{1}{c|}{} & \multicolumn{1}{c|}{}  
& \multicolumn{1}{c|}{\cellcolor[gray]{0.8}\textit{Weighted avg.}} & \cellcolor[gray]{0.8}\textit{0.967} & \cellcolor[gray]{0.8}\textit{{\bf 0.984}} & \cellcolor[gray]{0.8}\textit{{\bf 0.891}}  & \cellcolor[gray]{0.8}\textit{0.539} \\                                                 
\hline \hline
\multicolumn{1}{c|}{\multirow{8}{*}{\rotatebox[origin=c]{90}{\textbf{UNK21} $\sim$40,000 obs.}}}  
& \multicolumn{1}{c|}{\multirow{4}{*}{LR}}      
& \multicolumn{1}{c|}{Background} & 0.992 & 0.973 & 0.742  & 0.809  \\
\multicolumn{1}{c|}{} & \multicolumn{1}{c|}{}                         
& \multicolumn{1}{c|}{DoS}        & 0.928 & 0.962 & 0.789  & {\bf 0.606} \\
\multicolumn{1}{c|}{} & \multicolumn{1}{c|}{}  
& \multicolumn{1}{c|}{Port Scanning}        & 0.791 & 0.985 & 0.710  & 0.608 \\ 
\multicolumn{1}{c|}{} & \multicolumn{1}{c|}{}  
& \multicolumn{1}{c|}{\cellcolor[gray]{0.8}\textit{Weighted avg.}} & \cellcolor[gray]{0.8}\textit{0.943} & \cellcolor[gray]{0.8}\textit{0.973}
& \cellcolor[gray]{0.8}\textit{0.743}  & \cellcolor[gray]{0.8}\textit{{\bf 0.609}} \\ \cline{2-7} 
\multicolumn{1}{c|}{} 
& \multicolumn{1}{c|}{\multirow{4}{*}{RF}}      
& \multicolumn{1}{c|}{Background} & {\bf 0.994} & {\bf 0.987} & {\bf 0.831}  & 0.819 \\
\multicolumn{1}{c|}{} & \multicolumn{1}{c|}{}                         
& \multicolumn{1}{c|}{DoS}        & 0.928 & {\bf 0.985} & {\bf 0.877}  & 0.600 \\
\multicolumn{1}{c|}{} & \multicolumn{1}{c|}{}  
& \multicolumn{1}{c|}{Port Scanning}        & {\bf 0.799} & {\bf 0.991} & {\bf 0.780}  & 0.607 \\ 
\multicolumn{1}{c|}{} & \multicolumn{1}{c|}{}  
& \multicolumn{1}{c|}{\cellcolor[gray]{0.8}\textit{Weighted avg.}} & \cellcolor[gray]{0.8}\textit{{\bf 0.945}} & \cellcolor[gray]{0.8}\textit{0.987} 
& \cellcolor[gray]{0.8}\textit{0.830}  & \cellcolor[gray]{0.8}\textit{0.604} \\    

\bottomrule
\end{tabular}
\end{table*}

In order to verify the behaviour of these kind of ML-based solutions in generic network environments, another analysis across network datasets was performed. In this sense, Table \ref{tab:results_traintest_all} shows the AUC performance results when ML models (LR and RF) are fully trained using one of the previously mentioned datasets (train data column in the table) and fully tested on the other NIDS datasets (test data field in the table). This crossed analysis was carried out only for the common classes in the three datasets (i.e., \emph{Background}, \emph{DoS} and \emph{Port Scanning}). Gray cells in this table correspond to the performance presented in Tables~\ref{tab:results_cvugr16}-\ref{tab:results_cvkdd} ---i.e., analysis on one single dataset, which in the R-NIDS framework would occur when $x=1$ in Fig. \ref{fig:ngl-nids}---. Cells without numeric values means that no confident results were obtained mainly due to the low number of samples available to train the model for that specific class.

These results clearly expose the issues of ML-based NIDS solutions when they are built exclusively on one single dataset expecting them to perform well in generic network environments (i.e., in observations belonging to other network datasets not used to train the models). In fact, the performance of ML-based solutions drastically drops in this more realistic scenario, since the aim of this kind of solutions is to put them in production to detect possible attacks in generic network environments. In other words, we can safely conclude that the classic approach employed to build ML-based NIDS using one single dataset tend to overfit to observations of the network environment used to create that specific dataset, but results will very rarely generalize to other network environments. 

We computed some statistics on the results in Table \ref{tab:results_traintest_all} to support the previous conclusion. Particularly, the percentage of the mentioned accuracy loss of the models was calculated as an average for the three attacks. The highest loss was a $116.6\%$ decay in the performance of RF when trained with UGR'16 and tested on NSL-KDD, for 20,000 observations. In average, the models offered a $67.7\%$ accuracy loss when tested in other datasets different than the one used for training. The average accuracy of the models for all classes when trained and tested on the same dataset is $0.905$, while it worsens to $0.533$ when a different dataset is used for testing.

For a better understanding of these results, an additional combined Principal Component Analysis (PCA) study was carried out to unveil hidden behaviors of the involved classes and that, otherwise, it would be difficult to observe. The graphical analysis (score plots) performed on the derived datasets of 20,000 observations is shown in Fig.~\ref{fig:pca_analysis}. Since the equivalent analysis for the 10,000 observations version of the derived datasets is very similar, authors decided to omit it. Focusing on the three plots on the left of Fig.~\ref{fig:pca_analysis}, one can clearly distinguish two clusters independently of the class analyzed (\emph{Background}, \emph{DoS} or \emph{Port Scanning}). This result points out that UGR'16 observations are behaving completely different from the ones in the other two datasets. In fact, this also stands out in the results included in Table \ref{tab:results_traintest_all}, where the performance of the ML-based solutions trained on the UGR'16 dataset drastically drops when they are tested on any of the other two datasets. 

Moreover, on the three plots on the right of Fig.~\ref{fig:pca_analysis} we zoomed the graphical analysis of the corresponding score plots on the left side in order to show some details within the cluster. On these three plots, one can observe two clusters of slightly separated observations corresponding to the USNW-NB15 and NSL-KDD datasets. Besides, trend differences are also visible, specially for the \emph{Background} and \emph{Port Scanning} classes. However, this is not the case for the \emph{DoS} class where observations behave similarly in both datasets. Consequently, one could expect a better performance in the \emph{DoS} class than in the others when training or testing with USNW-NB15 and NSL-KDD datasets, fact that is, in general, observed in Table \ref{tab:results_traintest_all}. 

One last analysis was carried out on the UNK21 dataset in order to measure the benefits of using the R-NIDS framework when combining several network datasets. In this sense, the results included in column ``Overall'' in Table~\ref{tab:results_cvmix_zoom} show how ML-based models perform when using the whole UNK21 for training and testing. As it can be seen in the table, models achieve, on average, an AUC $\approx$ 0.8 for the \emph{Port Scanning} attack, AUC $\approx$ 0.95 for the \emph{DoS} attack and AUC $\approx$ 0.995 for the \emph{Background} traffic. In comparison  with the values in  Tables~\ref{tab:results_cvugr16}-\ref{tab:results_cvkdd}, the UNK21 slightly degrades the obtained performance for \textit{DoS} and \textit{Port Scanning} attacks in general. It can be motivated by the influence of the NSL-KDD dataset that pollutes, in some sense, the UNK21 dataset: almost the half part of the \textit{DoS} and \textit{Port Scanning} observations are added by the NLS-KDD (see Fig.~\ref{fig:classes_derived_datasets}). This fact, together with the nature of the NSL-KDD dataset, makes the models to perform slightly worse. Ne\-vertheless, the most interesting results arise when we focus on analysing the performance of these ML-based models on observations of only one of the network datasets, i.e., training them with the UNK21 dataset and testing only with samples corresponding to the specific network datasets used to build UNK21. These results can be seen in columns ``Only UGR'16'', ``Only UNSW-NB15'' and ``Only NSL-KDD'' in Table~\ref{tab:results_cvmix_zoom}. Overall, a high AUC of any of the ML models in these three columns would imply a better generalization, thus a better performance, on heterogeneous network environments. In concrete, the homogenization and aggregation of network datasets provided by the R-NIDS framework allowed to obtain reliable ML-based solutions that perform better in generic network environments than those built using the classic one-single-dataset approach. For instance, a RF model trained on the 20,000 observations version of the UNK21 dataset achieves an AUC of \{0.983, 0.892, 0.676\} for the \emph{Background} class in the UGR'16, UNSW-NB15 and NSL-KDD datasets, respectively. However, in Table~\ref{tab:results_traintest_all} the best results obtained for exactly the same setting were \{0.508, 0.814, 0.712\}, which is a significant improvement in two out of the three datasets. Similarly, a RF model trained on the 40,000 observations version of the UNK21 dataset achieves an AUC of \{0.991, 0.780, 0.607\} for the \emph{Port Scanning} class in the UGR'16, UNSW-NB15 and NSL-KDD datasets, respectively, while in Table~\ref{tab:results_traintest_all} the best results obtained for exactly the same setting were \{0.398, 0.562, 0.562\}, which in this case is a significant improvement in all the datasets.

This pattern is generally observed in Table~\ref{tab:results_cvmix_zoom} regardless the ML model and network dataset in which one wants to focus on. Therefore, the R-NIDS framework arises as a powerful tool to develop more robust and reliable ML-based NIDS for generic network environments. To the best of our knowledge, it is the first attempt that jointly considers several network datasets to build a ML-based NIDS and demonstrates the advantages of this approach, in terms of performance and generalization capabilities, in comparison to the classic one-single-dataset approach typically found in the literature.

\section{Conclusions and future work}
\label{sec:conclusions}

This work has presented a novel methodology for the design of Machine Learning (ML) based solutions for trustworthy and reliable Network Intrusion Detection Systems (NIDS), called Reliable-NIDS (or R-NIDS for short). In contrast to the existing methods, the ML models designed following the R-NIDS methodology achieve more accurate predictions in different network scenarios. Thus, they arise as more suitable tools for being deployed in a real environment, compared to previously existing ones. 

There is a large number of works in the literature addressing this problem with ML-based methods, and also several network datasets have been proposed. All these works use the same dataset for training and validating their mo\-dels, and unsurprisingly highly accurate results are published in most of the cases. However, this approach leads to models which overfit data from the given network environment used to train them, thus not being able to generalize to new data that can be present in generic scenarios scenarios or under different conditions. Therefore, these existing methods are not suitable to be deployed in real environments. 

Through an extensive experimentation, this work has demonstrated how two well-known ML models (a linear and a non-linear ones), commonly used for NIDS, suffer from the mentioned overfitting problem when trained on one network dataset and validated on a different one. Indeed, the obtained performance loss was up to $116.6\%$ when validating the model on a different dataset. For this study, we chose three well accepted datasets, namely UGR'16, USNW-NB15, and NSL-KDD. Although these datasets have different features and making a joint analysis was a priory unfeasible, it turned out to be possible thanks to the use of the FaaC feature engineering method present in R-NIDS. This technique makes the homogenization of different network datasets possible, creating different derived datasets from them (all having the same variables), thus being a key driver to perform data aggregation over several heterogeneous datasets.

Additionally, this work has introduced a novel dataset, called UNK21. It was built from the samples contained in three of the most studied datasets in the literature (UGR'16, USNW-NB15, and NSL-KDD), that could be integrated using the R-NIDS methodology proposed. Last but not least, by using the UNK21 dataset this paper has shown how the two ML models considered in this work were able to avoid overfitting, thus leading to more reliable ML-based NIDS solutions compared to those trained on just one single network dataset.

Finally, this work opens interesting research lines in the field. In general, the authors are interested in extending the study with other state-of-the-art ML methods from the literature, as well as with other existing relevant datasets. Besides, and given the importance of integrating different datasets for building generic methods as it has been shown in this paper, the authors consider interesting to work towards the design and exploration of other generalization mechanisms based on different ways of integrating the main network datasets in the field with minimum information loss.


%

\ifCLASSOPTIONcompsoc
  \section*{Acknowledgments}
\else
  \section*{Acknowledgment}
\fi

This work was supported by the Spanish Ministerio de Ciencia, Innovaci\'on y Universidades and the ERDF under contracts RTI2018-100754-B-I00 (iSUN) and RTI2018-098160-B-I00 (DEEPAPFORE), ERDF under project FEDER-UCA18-108393 (OPTIMALE), and Junta de Andaluc\'ia and ERDF (GENIUS -- P18-2399).

\ifCLASSOPTIONcaptionsoff
  \newpage
\fi



\bibliographystyle{IEEEtran}
%

\bibliography{refs.bib}

%

\begin{IEEEbiography}[{\includegraphics[width=1in,height=1.25in,clip,keepaspectratio]{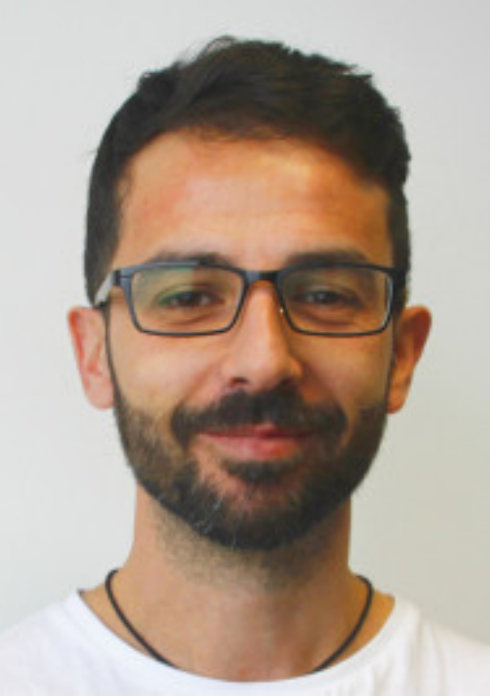}}]{Roberto Magán-Carrión}
is an Assistant Professor at the Department of Signal Theory, Telematics and Communications at the University of Granada (Spain) and member of “Network Engineering and Security Group  (NESG)” research group. He received his Ph.D. in ICT in 2016 at the University of Granada with a Cum Laude grade. He also worked as a post-doctoral fellow at the University of Cádiz, as part of a research talent attraction international program promoted by this university. His research interests are focused on security in heterogeneous communications networks and systems, specifically on anomaly detection and classification, response and resilient solutions.
\end{IEEEbiography}

\begin{IEEEbiography}[{\includegraphics[width=1in,height=1.25in,clip,keepaspectratio]{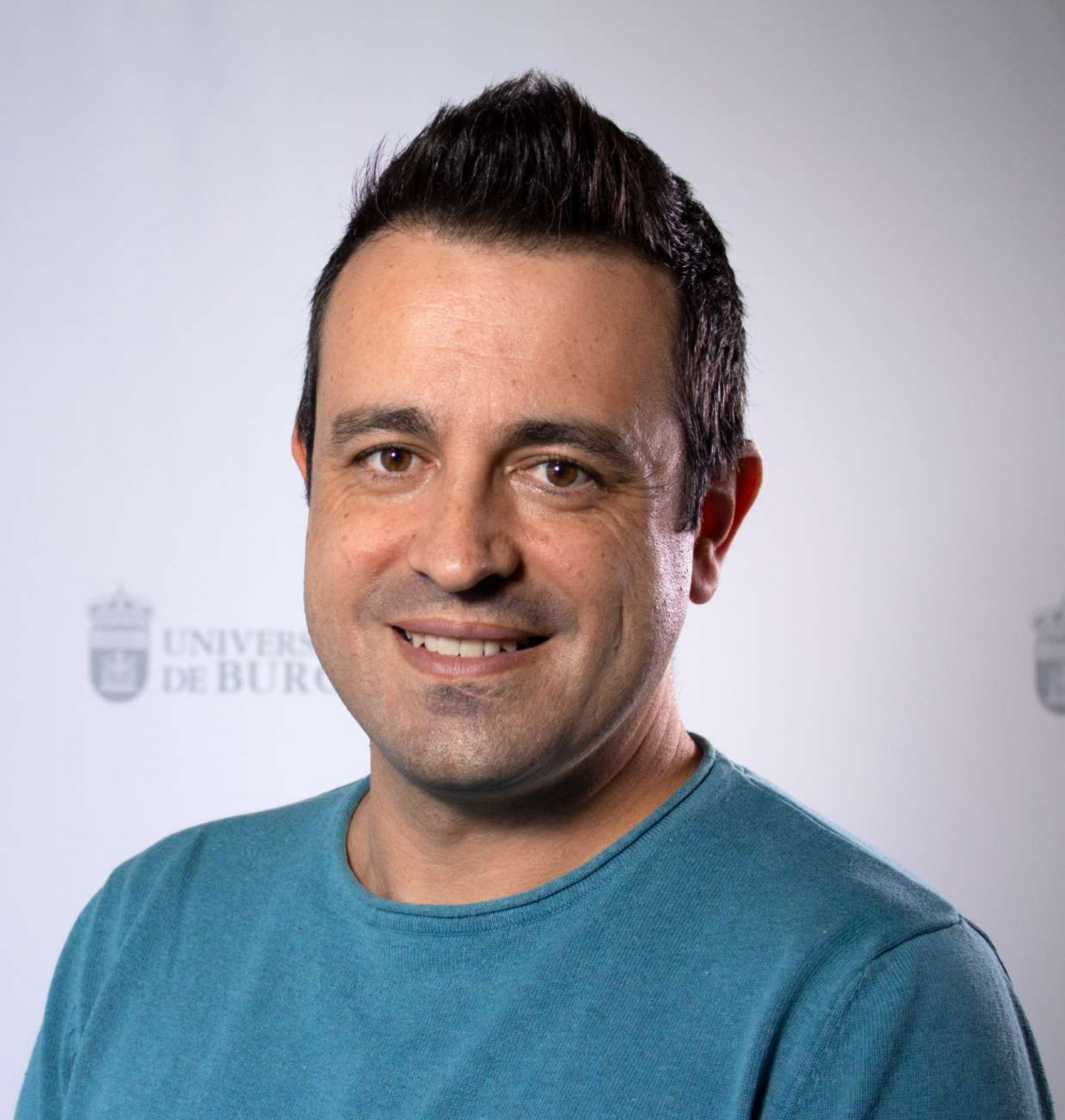}}]{Daniel Urda}
holds a Ph.D. in Computer Science (2015) from the University of Málaga (Spain) and is currently an Assistant Professor of the department of Computer Science at the University of Burgos (Spain). His research mainly involves the development and application of machine learning methods for several domains such as Bioinformatics, Cybersecurity, Air pollution, Logistics \& Transportation and Industry. He has a strong publication record in relevant journals and international conferences ($>$50 papers) and has been guest editor of several special issues in well-known journals related to Computer Science. At the university, he is responsible for teaching several courses of Bachelors and Masters degree in Computer Science such as Artificial Intelligence or Neural Computing, among others.
\end{IEEEbiography}

\begin{IEEEbiography}[{\includegraphics[width=1in,height=1.25in,clip,keepaspectratio]{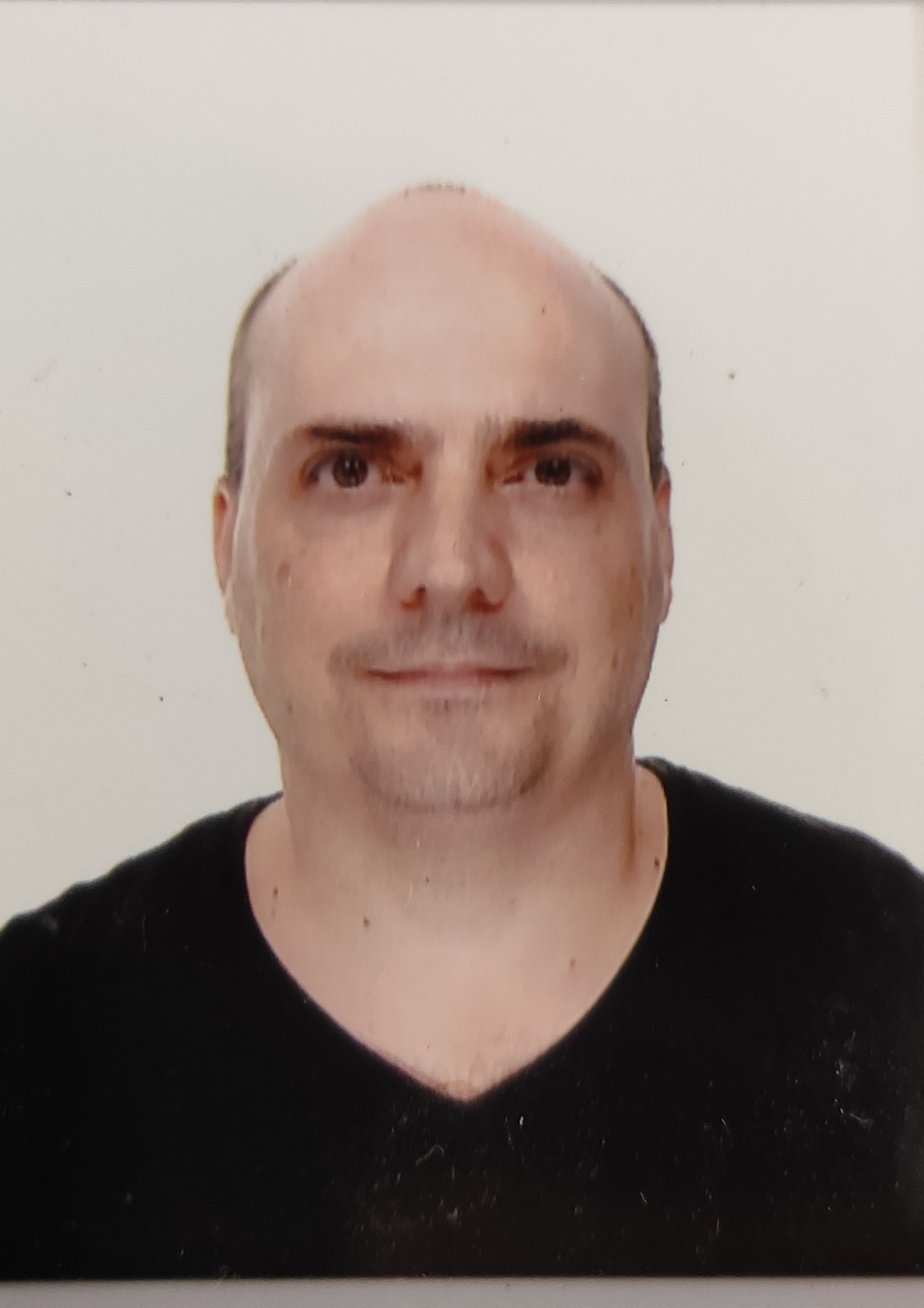}}]{Ignacio Díaz-Cano}
is a Predoctoral Researcher at the Dpt. of Automatic, Electronic, Computer Architecture \& Communication Networks
Engineering from the University of Cádiz (Spain). His research mainly involves two main areas of knowledge: cybersecurity and industrial robotics, with some publications and conferences in both areas.
\end{IEEEbiography}

\begin{IEEEbiography}[{\includegraphics[width=1in,height=1.25in,clip,keepaspectratio]{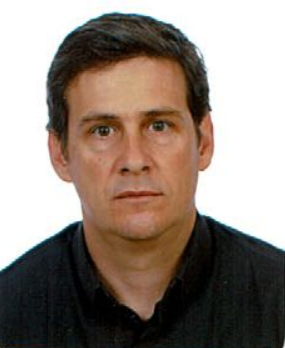}}]{Bernabé Dorronsoro} holds a Ph.D. in Computer Science (2007) from the University of Málaga, Spain. He worked for more than seven years in Luxembourg and Lille universities, and he currently serves as an associate professor at the University of Cádiz (Spain), where he leads the GOAL research group (\url{https://goal.uca.es}). His main research interests include green computing, sustainable transportation, Grid and Cloud computing, Intelligent Transportation Systems, complex problems optimization, and machine learning. He has published over 50 journal papers and five books. Dr Dorronsoro is associate editor of the International Journal of Metaheuristics, and member of the editorial board of several journals: Engineering Applications of Artificial Intelligence, Applied Sciences, International Journal of High Performance Systems Architecture, International Journal of Innovative Computing and Applications, and Progress in Artificial Intelligence.
\end{IEEEbiography}







\end{document}